\documentclass{article} %
\usepackage{iclr2027_conference,times}

\usepackage{amsmath,amsfonts,bm}

\def\eqref#1{equation~\ref{#1}}

\def\1{\bm{1}}

\DeclareMathAlphabet{\mathsfit}{\encodingdefault}{\sfdefault}{m}{sl}
\SetMathAlphabet{\mathsfit}{bold}{\encodingdefault}{\sfdefault}{bx}{n}

\usepackage{pifont}
\usepackage{graphicx}
\usepackage{float}
\usepackage{booktabs}
\usepackage{multirow}
\usepackage{bbm}
\usepackage[table]{xcolor}
\definecolor{ciband}{RGB}{224,235,247} %
\definecolor{cdband}{RGB}{251,233,222} %
\definecolor{cicol}{RGB}{31,101,178}
\definecolor{cdcol}{RGB}{193,94,31}
\newcommand{\catCI}{\textcolor{cicol}{\textbf{CI}}}
\newcommand{\catCD}{\textcolor{cdcol}{\textbf{CD}}}
\newcommand{\ctype}[1]{\textsf{\small #1}}
\newcommand{\cdg}{\ensuremath{\mathrm{CD}_{g}}}
\newcommand{\milag}{\ensuremath{\mathrm{MI}_{\mathrm{lag}}}}
\usepackage{subcaption}
\usepackage{flafter} %
\usepackage{hyperref}
\usepackage{url}
\usepackage[textsize=small]{todonotes}
\setuptodonotes{inline}

\title{MixBench-TS: A Multivariate Time Series \\ Forecasting Benchmark \\ Where Channel Mixing Pays Off}

\author{Ibram Abdelmalak, Mischa Putzke, Jungmin Choi, Tom Hanika, \\
\textbf{Vijaya Krishna Yalavarthi \& Lars Schmidt-Thieme} \\
Information Systems and Machine Learning Lab (ISMLL)\\
University of Hildesheim \\
Hildesheim, 31141, Germany \\
{\small\texttt{\{abdelmalak, putzke, choi, hanika, yalavarthi, schmidt-thieme\}@ismll.de}}
}

\iclrfinalcopy %
\begin{document}

\maketitle
\lhead{Preprint. Under review.}

\begin{abstract}
Multivariate Time Series Forecasting (MTSF) models that mix information across channels assume that the past of one channel carries information about the future of another. Yet they are evaluated on a small fixed set of standard datasets whose cross-channel structure is rarely examined. We ask two questions: \emph{How can we reliably measure lagged, non-linear, and joint coupling in MTSF datasets?} and \emph{Do the standard datasets actually have such coupling?} To answer the first, we test four candidate measures on synthetic datasets with planted ground-truth coupling: Granger Causality (GC), Transfer Entropy (TE), lagged Mutual Information (MI), and the CD gain, a model-based measure we introduce that compares a channel-dependent (CD) model to its channel-independent (CI) variant. Only lagged MI and the CD gain recover every planted coupling. For the second question, the answer is a definite \emph{no}, as the standard datasets have a median of only 23\% lagged-coupled channel pairs and a median CD gain of $-4.9\%$, compared to 78\% and $+4.6\%$ on chaotic ODE systems. We therefore propose \textbf{MixBench-TS}, a benchmark of 10 real-world datasets with a median of 55.5\% lagged-coupled pairs and a median CD gain of $+1.7\%$. Across six state-of-the-art models tuned under one protocol, CI models win on 10/10 (MSE) and 8/10 (MAE) standard datasets, but on only 3/10 and 2/10 MixBench-TS datasets. We recommend using our benchmark for evaluating new CD models. Moreover, we propose profiling new datasets with lagged MI and the CD gain before using them to evaluate multivariate models. Code and data are available at \url{https://anonymous.4open.science/r/mixbench-ts-B027}.
\end{abstract}

\section{Introduction}

\begin{figure}[t]
\centering
\begin{minipage}[t]{0.49\linewidth}\vspace{0pt}\centering
  \includegraphics[width=\linewidth]{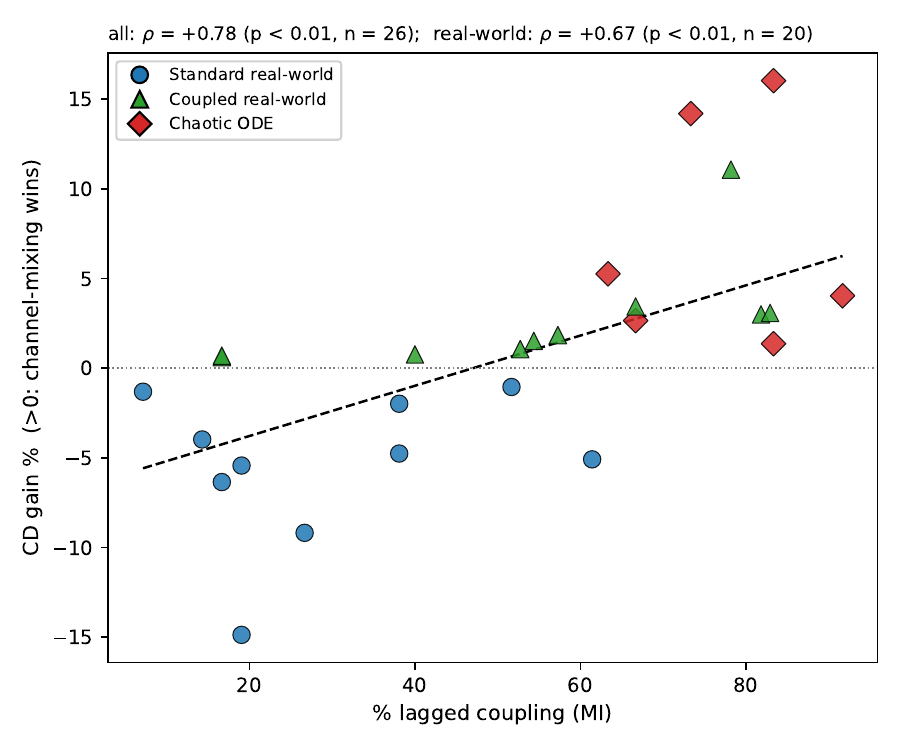}
  \caption{Lagged coupling vs.\ channel-mixing payoff on 26 datasets (10 standard, 10 coupled, 6 ODE). $x$: lagged MI (\milag{}), the share of channel pairs coupled more strongly at a lag than at lag 0. $y$: CD gain (\cdg{}), the MSE reduction of TSMixer over TSMixer\textsubscript{CI} ($>0$: mixing helps). Dashed: linear fit. Spearman correlations above the plot.}
  \label{fig:motivation}
\end{minipage}\hfill
\begin{minipage}[t]{0.49\linewidth}\vspace{0pt}\centering\footnotesize\setlength{\tabcolsep}{2.5pt}
\providecommand{\cmark}{\ding{51}}\providecommand{\xmark}{\ding{55}}
  \captionof{table}{Retrieval of the coupling planted into the target $c_0$ of each synthetic dataset. \cmark{}: exactly the planted source(s), at the planted lag where the measure resolves lags. \xmark{}: missed, mis-located or false. \cdg{} uses identical hyperparameters for both variants and the \ctype{independent} dataset as reference (3 seeds). \textbf{Bold}: all retrieved. Details in Table~\ref{tab:synth_retrieval_detailed}.}
  \label{tab:synth_retrieval}
  \vspace{4pt}
\newcolumntype{Y}{>{\centering\arraybackslash}p{0.85cm}}
\begin{tabular}{l YYYY}
\toprule
Dataset & \cdg{} & \milag{} & TE & GC \\
\midrule
\ctype{independent} & \cmark & \cmark & \cmark & \cmark \\
\ctype{contemp-linear} & \cmark & \cmark & \xmark & \cmark \\
\ctype{lagged-linear} & \cmark & \cmark & \cmark & \cmark \\
\ctype{lagged-non-linear} & \cmark & \cmark & \cmark & \xmark \\
\ctype{multiplicative-joint} & \cmark & \cmark & \cmark & \xmark \\
\ctype{additive-mixed} & \cmark & \cmark & \cmark & \xmark \\
\midrule
Retrieved & \textbf{6/6} & \textbf{6/6} & 5/6 & 3/6 \\
\bottomrule
\end{tabular}
\end{minipage}
\end{figure}

Multivariate Time Series Forecasting (MTSF) models that mix information across channels are designed on an assumption of lagged coupling: \emph{a source channel's past carries additional information about a target channel's future}. Contemporaneous coupling alone is not enough, since the source's value at the timestep of forecasting is unknown as well. On datasets whose cross-channel coupling is weak or mostly contemporaneous, a channel-dependent (CD) model therefore gains nothing over a channel-independent (CI) model, which forecasts each channel from its own past. Such datasets are a not a good fit for evaluating CD models.

Prior work has characterized MTSF datasets along axes such as domain coverage and temporal properties \citep{godahewa2021monash,qiu2024tfb}, but not along the lagged coupling that CD models rely on. The closest effort \citep{abdelmalak2026channel} uses ODE datasets from \citet{gilpin2021chaos} to argue that current datasets lack channel coupling. However, it relies on Granger Causality (GC)\citep{granger1969investigating}, which cannot capture non-linear lagged coupling. Another limitation is that they consider only the standard and ODE datasets. This leaves two open questions:
\begin{itemize}
	\item[Q1.] \emph{How can we reliably measure lagged, non-linear, and joint coupling in MTSF datasets?}
	\item[Q2.] \emph{Once measured, do the standard datasets used by the field \citep{lai2018modeling,zhou2021informer,wu2021autoformer} actually have such coupling?}
\end{itemize}

To address Q1, we curate six synthetic datasets with known planted couplings. On this suite, we test four candidate measures: Granger Causality (GC; \citealp{granger1969investigating}), Transfer Entropy (TE; \citealp{schreiber2000measuring}), lagged Mutual Information (\milag{}; \citealp{shannon1948mathematical,cover2006elements}), and the CD gain (\cdg{}), a model-based measure we introduce that compares a CD forecaster to its CI variant, following the landmarking concept of \citet{pfahringer2000meta}. GC misses the non-linear and joint couplings, and TE places the contemporaneous coupling at a wrong, positive lag. Only \milag{} and \cdg{} recover every planted coupling (Table~\ref{tab:synth_retrieval}), so we adopt them as our two profiling tools.

We then address Q2 by measuring \milag{} and \cdg{} on the 10 standard datasets, using six datasets from the chaotic ODE benchmark \citep{gilpin2021chaos} as a known-correlated reference. The answer is a \emph{definite no}, as the standard datasets show weak lagged coupling compared to the ODE datasets (median of 23\% vs.\ 78\% lagged-coupled pairs, Figure~\ref{fig:motivation} and Table~\ref{tab:analysis_standard}). Moreover, a CI variant of TSMixer beats the full channel-mixing model on all 10 standard datasets, while channel mixing helps on every ODE dataset (median \cdg{} of $-4.9\%$ vs.\ $+4.6\%$).

If the standard datasets fall short, the natural next step is to find datasets that have the lagged coupling that channel mixing needs. We therefore propose \textbf{MixBench-TS}, a benchmark of 10 real-world datasets that exist in the literature but are rarely used for MTSF evaluation \citep{qiu2024tfb,qiao2026its,stein2025causalrivers,baigutanova2025continuous}, selected by their positive \cdg{}. They cover environmental monitoring, energy, hydrology, marine sensing, and health, and we refer to them throughout as \emph{the coupled datasets}. They show substantially stronger lagged coupling than the standard datasets.

To test whether this pattern carries over to actual forecasting, we evaluate six state-of-the-art models spanning both channel strategies (CI and CD), tuned under one unified protocol that also covers the lookback window \citep{abdelmalak2026channel}. On the coupled datasets, where lagged coupling is substantially stronger, CI models win on only 3 of 10 (MSE), compared to all 10 standard datasets.

Our contributions are as follows:
\begin{itemize}
	\item We introduce the CD gain (\cdg{}), a model-based measure of exploitable cross-channel coupling in MTSF datasets, and validate it alongside three other candidate measures on synthetic datasets with known ground-truth coupling (Sections~\ref{sec:cd-gain} and~\ref{sec:measurement-evaluation}).
	\item We show that the datasets the field treats as standard are poor tests of CD models (Section~\ref{sec:std-measurement}).
	\item We propose MixBench-TS, a benchmark of 10 real-world coupled datasets that are rarely used for MTSF evaluation and carry substantially stronger lagged coupling than the standard datasets (Section~\ref{sec:new-datasets}).
	\item We benchmark six models under a unified tuning protocol and show that the CI advantage collapses on the coupled datasets (Section~\ref{sec:new-benchmark-results}).\footnote{Code and data, including the profiling toolkit (\milag{}, \cdg{}) and the synthetic data generator, are available at \url{https://anonymous.4open.science/r/mixbench-ts-B027}.}
\end{itemize}

\section{Related Work}
\label{sec:related-work}

\paragraph{Channel mixing strategies in MTSF models.} There has been rising interest in building MTSF models with varying architectures. We categorize recent models into CI and CD models based on whether a model treats each channel as a separate time series or mixes information across channels. The CI category covers linear \citep{zeng2023transformers}, MLP-based \citep{lin2024cyclenet}, and even transformer-based models that still operate on each channel in isolation \citep{nie2022time}. The CD category includes transformer-based models that apply attention across channels \citep{zhang2023crossformer, liu2024itransformer, chen2025simpletm} and MLP-based or linear models that mix channels \citep{chen2023tsmixer, yue2026olinear}. It also covers CNN-based models \citep{luo2024moderntcn} and frequency-domain-based methods \citep{wang2024timemixer, wang2025timemixer++}. More recent CD models mix channels selectively rather than fully, using graph-based filtering \citep{hu2025timefilter} or channel clustering \citep{qiu2025duet, chen2024similarity}. 

\paragraph{Benchmarks and evaluation.} Many recent papers have extended the datasets available for MTSF in different directions, such as ODE-based benchmarks \citep{gilpin2021chaos, klotergens2025physiome}, bursty network-traffic benchmarks \citep{guthula2026netburst}, causality-driven benchmarks \citep{stein2025causalrivers}, and application-traffic benchmarks balanced across forecasting regimes \citep{xue2026quitobench}. Other benchmarks have been proposed to extend domain coverage and characterize temporal properties of the data \citep{qiu2024tfb, godahewa2021monash, aksu2025gifteval}. On the \emph{evaluation} side, \citet{zeng2023transformers} show that a simple linear model outperforms transformers on the standard datasets. Meanwhile, \citet{chen2025closer} analyze the existing benchmark based on transformer-based models' performance. \citet{roque2025cherry} investigate how papers choose datasets to evaluate on, while \citet{brigato2025there} study the effect of hyperparameter tuning and forecasting horizons on model performance on the standard datasets. Closer to our work, some studies investigate when cross-channel modeling pays off. For instance, \citet{han2024capacity} attribute the strong performance of CI models to their robustness to distribution shift. \citet{shao2025exploring} link model performance to the level of heterogeneity within datasets. Finally, \citet{abdelmalak2026channel} show via Granger Causality that the standard datasets carry little linear channel coupling. We build on this literature by following rigorous evaluation practices in our protocol, and by extending the analysis to non-linear and lagged coupling.

\section{Problem Formulation}

Each dataset is a multivariate time series $\mathbf{Z} = (\mathbf{z}_1, \dots, \mathbf{z}_T) \in \mathbb{R}^{T \times C}$ with $C$ channels (variates) observed at $T$ equally spaced steps. Following \citet{qiu2024tfb}, $\mathbf{Z}$ is split \emph{chronologically} into training, validation, and test segments, and samples are built with a sliding window of stride 1. A model $f_{\boldsymbol{\theta}} : \mathbb{R}^{L \times C} \to \mathbb{R}^{H \times C}$ maps a \emph{lookback window} of the last $L$ steps to the next $H$ steps (the \emph{forecast horizon}), jointly over all $C$ channels. Forecast targets never cross a split boundary (Appendix~\ref{sec:problem-formulation-detailed}). The standard datasets commonly use $H \in \{96, 192, 336, 720\}$, while for the remaining datasets we adapt the horizons to be semantically relevant (Appendix~\ref{sec:datasets-details}).

\section{Channel Dependency Measurement}
\label{sec:statistical-profiling}
We consider four complementary measures of cross-channel coupling, which differ in the type of coupling they target and in how they are computed. Granger Causality (GC), Transfer Entropy (TE), and Lagged Mutual Information (\milag{}) are statistical and are computed \emph{pairwise and directionally} on the data itself, whereas the CD gain (\cdg{}) is model-based. Each pairwise measure is evaluated on every ordered source--target pair of the channel set $\mathcal{C}$,
\begin{equation}
\mathcal{P}=\{(x,y): x,y\in\mathcal{C},\, x\neq y\},\qquad |\mathcal{P}|=C(C-1),
\end{equation}
and then aggregated to a single dataset value. On wide datasets, $|\mathcal{C}|$ is capped at $40$ by uniform random sampling of channels (sensitivity in Appendix~\ref{sec:parameter-sensitivity}). Further implementation details are given in Appendix~\ref{sec:stats-profiling-implementation-details}.

\paragraph{Granger Causality (GC).}
\label{sec:gc}
Each pair is jointly made stationary (seasonal and ADF-selected regular differencing, \citealp{said1984testing}), so GC reflects directed predictive structure rather than shared trend or seasonality. We then fit two OLS regressions with $p$ lags, where $p$ is the smallest candidate lookback window of the dataset, on the $N$ usable time steps: a univariate \emph{restricted} model that predicts $y_t$ from its own past $\{y_{t-1},\dots,y_{t-p}\}$, and a multivariate model that adds the source's past $\{x_{t-1},\dots,x_{t-p}\}$ \citep{granger1969investigating}. With $\mathrm{SSR}_{\text{u}}, \mathrm{SSR}_{\text{mv}}$ their sums of squared residuals and $\nu_{\text{u}}=N-(p{+}1)$, $\nu_{\text{mv}}=N-(2p{+}1)$ their residual degrees of freedom, our metric is the degrees-of-freedom corrected partial $R^2$ (Geweke linear feedback, \citealp{geweke1982measurement}), taken as the median over pairs:
\begin{equation}
R^2_{\mathrm{GC}}(x\!\to\!y)=1-\frac{\mathrm{SSR}_{\text{mv}}/\nu_{\text{mv}}}
{\mathrm{SSR}_{\text{u}}/\nu_{\text{u}}},
\qquad
\widetilde{R^2_{\mathrm{GC}}}=\operatorname*{median}_{(x,y)\in\mathcal{P}} R^2_{\mathrm{GC}}(x\!\to\!y).
\end{equation}
Intuitively, $R^2_{\mathrm{GC}}$ is the fraction of $y$'s residual variance that $x$'s past removes. It is $\approx 0$ under the null of no Granger causality and moves toward $1$ the more $x$'s past helps predict $y$. Dividing by the residual degrees of freedom cancels the mechanical decrease of $\mathrm{SSR}$ from the $p$ extra regressors, which centers the null near $0$ regardless of $p$ or $N$.

\paragraph{Transfer Entropy (TE).}
\label{sec:te}
TE uses the same joint differencing as GC, so the linear and non-linear measures see identical series. It is the directed, non-linear generalization of GC, approximated by the rank-space conditional mutual information $I\!\left(y_t;x_{t-\ell}\mid y_{t-1}\right)$ \citep{schreiber2000measuring,kraskov2004estimating}. Here, $y_t$ is the target's present, $x_{t-\ell}$ the source's value $\ell$ steps earlier, and $y_{t-1}$ the target's most recent past. As in the restricted GC model, conditioning on $y_{t-1}$ removes what the target's own past already explains, so only the \emph{additional} information carried by the source is counted. We sweep the lags $\ell\in\{1,\dots,p\}$, with $p$ as for GC, and select $\ell^\star$ as the statistically significant lag with the largest effective value (Appendix~\ref{sec:stats-profiling-implementation-details}). Our metric maps the effective TE onto the $R^2$ scale of GC and, as for GC, takes the median over pairs:
\begin{equation}
R^2_{\mathrm{TE}}(x\!\to\!y)=1-e^{-2\,\mathrm{TE}^{\mathrm{eff}}(x\!\to\!y)},
\qquad
\widetilde{R^2_{\mathrm{TE}}}=\operatorname*{median}_{(x,y)\in\mathcal{P}}
R^2_{\mathrm{TE}}(x\!\to\!y),
\end{equation}
where $\mathrm{TE}^{\mathrm{eff}}(x\!\to\!y)=\max\!\big(0,\,
\widehat{I}(y_t;x_{t-\ell^\star}\mid y_{t-1})-\overline{I}_{\text{null}}\big)$ is the debiased TE \citep{marschinski2002analysing}, with $\overline{I}_{\text{null}}$ the mean of a permutation-based surrogate null \citep{runge2018conditional}. The mapping $R^2=1-e^{-2I}$ is the Gaussian information--$R^2$ identity.

\paragraph{Lagged Mutual Information (\milag{}).}
\label{sec:mi}
Unlike GC and TE, MI does not condition on the target's own past. At lag $0$, the MI between $x_t$ and $y_t$ is a \emph{symmetric}, common-driver measure of their coupling. At a positive lag, the MI between $x_{t-\ell^\star}$ and $y_t$ is \emph{directional}, from source to target, as in GC and TE. We compute it on the \emph{raw} series (rank-transformed, no differencing) with a KSG $k$-NN estimator \citep{kraskov2004estimating}, at lag $0$ and at the positive lag $\ell^\star$ that maximizes the effective MI over the geometric lag grid $\Lambda$ (Eq.~\ref{eq:mi-lags}, Appendix~\ref{sec:stats-profiling-implementation-details}). Our metric, the \emph{lagged-coupling fraction} $\phi_{\mathrm{lag}}$, summarizes whether lagged or contemporaneous coupling dominates,
\begin{equation}
\phi_{\mathrm{lag}}=\frac{1}{|\mathcal{P}|}\sum_{(x,y)\in\mathcal{P}}
\mathbbm{1}\!\left[\,\widehat{I}\!\left(x_{t-\ell^\star};y_t\right)>
\widehat{I}\!\left(x_t;y_t\right)\right],
\end{equation}
where $\widehat{I}(a;b)$ is the effective (surrogate-debiased) KSG estimate of the MI between $a$ and $b$, and $\mathbbm{1}[\,\cdot\,]$ is the indicator function. $\phi_{\mathrm{lag}}$ is thus the share of pairs whose lagged coupling exceeds their contemporaneous coupling (reported in percent in Tables~\ref{tab:analysis_standard} and~\ref{tab:occasional_results}). A high value means cross-channel information is predominantly \emph{lagged} (delayed transfer) rather than \emph{contemporaneous} (a shared instantaneous driver). This matters because contemporaneous coupling often carries little useful information for forecasting, whereas lagged coupling can supply the extra predictive signal that a multivariate model exploits. Throughout the paper, we refer to this measure as \milag{} and report $\phi_{\mathrm{lag}}$ as its dataset-level value.

\paragraph{CD Gain (\cdg{}).}
\label{sec:cd-gain}
The statistical measures above test for specific forms of coupling (linear, non-linear, lagged) between pairs of channels, and can therefore miss coupling that only emerges jointly across several channels. We therefore follow the landmarking concept of \citet{pfahringer2000meta}, which characterizes a dataset through the performance of simple learners on it. We train two TSMixer variants \citep{chen2023tsmixer} that differ only in their MLP-based channel-mixing component: TSMixer, which mixes information across channels, and TSMixer\textsubscript{CI}, which drops the channel-mixing layers and forecasts each channel independently. \cdg{} is the relative reduction in test MSE obtained through channel mixing,
\begin{equation}
\cdg = 100 \cdot \frac{\mathrm{MSE}_{\mathrm{CI}} - \mathrm{MSE}_{\mathrm{CD}}}{\mathrm{MSE}_{\mathrm{CI}}},
\label{eq:cd-gain}
\end{equation}
where $\mathrm{MSE}_{\mathrm{CI}}$ and $\mathrm{MSE}_{\mathrm{CD}}$ denote the test MSE of TSMixer\textsubscript{CI} and TSMixer, averaged over the forecast horizons and three random seeds. A positive \cdg{} indicates that a multivariate model can exploit the cross-channel information, whereas a non-positive value indicates that channel mixing brings no benefit or even hurts the forecasting performance.

\section{Evaluation of Channel Dependency Measurements}
\label{sec:measurement-evaluation}

\subsection{Synthetic Datasets Curation}
\label{sec:synthetic-datasets-curation}

To check whether the four measures capture couplings of varying types, we build synthetic datasets with known ground-truth coupling, which allows us to verify whether each measure recovers it. The datasets span three classes of cross-channel coupling: (i) no coupling, (ii) pairwise coupling, where a single source channel drives a target channel, either contemporaneously or with a lag and either linearly or non-linearly, and (iii) joint coupling, where several source channels drive the same target channel. Each base channel $c_i$ is an Autoregressive (AR) process of order $p_i$ driven by independent Gaussian noise, plus a seasonal component with a channel-specific period. We then plant a coupling from one or more source channels into a target channel $c_0$, according to its coupling type: $\mathit{coupling} \in \{$\ctype{contemp-linear}, \ctype{lagged-linear}, \ctype{lagged-non-linear}, \ctype{multiplicative-joint}, \ctype{additive-mixed}$\}$.

\begin{figure}[t]
\centering
\includegraphics[width=\linewidth]{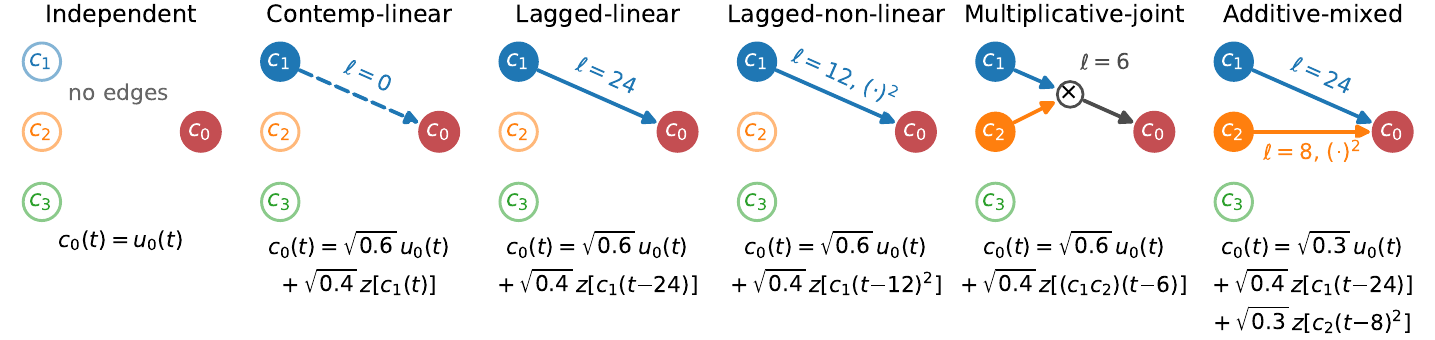}
\caption{Planted coupling structure of the six synthetic datasets. Arrows mark the couplings into the target $c_0$, labeled with lag $\ell$ and form (dashed: contemporaneous, $(\cdot)^2$: squared, $\times$: multiplicative). Hollow nodes have no edge into $c_0$. Below each panel is the generating equation of $c_0$, with $u_0$ its standardized base process, $z[\cdot]$ standardization, and squared weights as variance shares (Appendix~\ref{sec:synthetic-data-generation-details}).}
\label{fig:synth_graphs}
\end{figure}

\begin{figure}[t]
\centering
\includegraphics[width=0.85\linewidth]{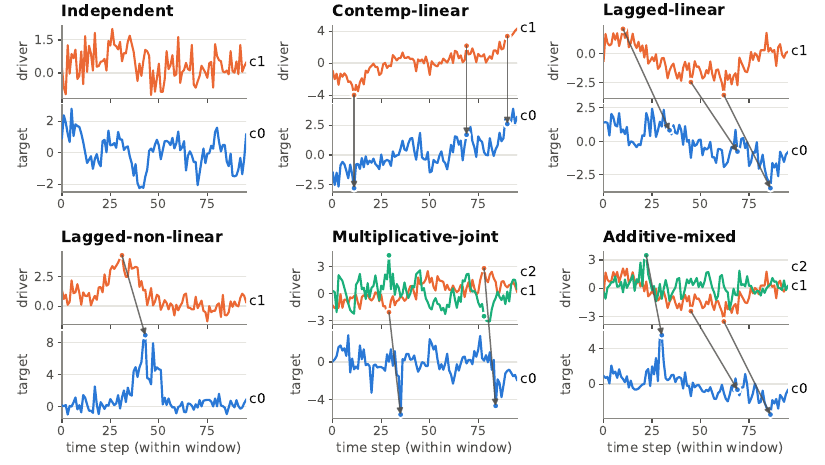}
\caption{Planted couplings on a short window of each synthetic dataset, with the source channel(s) (labeled driver) on top and the target $c_0$ below. Each arrow links a source value to its imprint on $c_0$ after the planted lag $\ell$. This is $\ell=0$ (vertical) for \ctype{contemp-linear}, and $24$, $12$ and $6$ for \ctype{lagged-linear}, \ctype{lagged-non-linear} and \ctype{multiplicative-joint}, respectively. In \ctype{additive-mixed}, $c_2$ arrives after $\ell=8$ and $c_1$ after $\ell=24$. \ctype{independent} has no coupling and no arrows.}
\label{fig:synth_planted_grid}
\end{figure}

Figure~\ref{fig:synth_graphs} shows the six datasets and their planted coupling structure. The \ctype{independent} dataset establishes the null values of the measures. The three pairwise datasets couple a single source $c_1$ into $c_0$, and the two joint datasets couple $c_1$ and $c_2$ into $c_0$. In \ctype{multiplicative-joint}, the target depends on the product of both sources, so neither source affects it on its own. Figure~\ref{fig:synth_planted_grid} visualizes these couplings, where each arrow links a source value to its imprint on $c_0$ at the planted lag. Each dataset has $C=4$ channels and $T=20{,}000$ timesteps (details in Appendix~\ref{sec:synthetic-data-generation-details}).

\subsection{Evaluation of Measurements on Synthetic Datasets}
\label{sec:statistical-methods-validation}
We summarize whether each of the four measures retrieves the planted coupling of the six datasets, with the underlying values in Table~\ref{tab:synth_retrieval_detailed}. For the three pairwise measures, let $D_s(\ell)$ be the lag profile of the measure from source $s$ into the target $c_0$ over the lag grid $\Lambda_{\mathrm{syn}}=\{1,\dots,48\}$ (plus lag $0$ for \milag{}), and $D^{(b)}_s(\ell)$ its value on the $b$-th of $B$ circular-shift surrogates. We compute
\begin{align}
\tau_{95} = Q_{0.95}\Bigl(\bigl\{\max_{s,\;\ell\in\Lambda_{\mathrm{syn}}} D^{(b)}_s(\ell)\bigr\}_{b=1}^{B}\Bigr),
\qquad
r_s &= \frac{\max_{\ell\in\Lambda_{\mathrm{syn}}} D_s(\ell)}{\tau_{95}},
\label{eq:synth-retrieval}\\
\hat\ell_s &= \arg\max_{\ell\in\Lambda_{\mathrm{syn}}} D_s(\ell). \nonumber
\end{align}
Here, $Q_{0.95}(\cdot)$ is the empirical 95\% quantile over the $B$ surrogates. Taking the maximum over all sources and lags corrects for testing many source--lag combinations, so that without coupling, any source exceeds $\tau_{95}$ at any lag with only 5\% probability. A source is flagged if $r_s>1$, and $\hat\ell_s$ is its detected lag. GC is cumulative over VAR orders and therefore gives no lag. A dataset counts as retrieved only if exactly the planted source(s) are flagged at the planted lag. For \cdg{}, we compare the gain of TSMixer over TSMixer\textsubscript{CI} on each coupled dataset to that on the \ctype{independent} reference, and count a dataset as retrieved if this excess gain is significantly positive over 3 seeds.

All four measures correctly report no coupling on \ctype{independent} and retrieve the \ctype{lagged-linear} coupling, but differ once the coupling is contemporaneous, non-linear, or joint. GC, being linear and conditioned on the source's past, flags the contemporaneous coupling only weakly ($r=1.5$). It stays below the null on the non-linear coupling and on the squared source $c_2$ of \ctype{additive-mixed}, and barely exceeds it on the multiplicative coupling ($r=1.1$ for $c_1$, below the null for $c_2$). TE retrieves all lagged couplings but, restricted to only positive lags, mis-locates the \ctype{contemp-linear} coupling at lag $24$ ($r=11$). Only \milag{} and \cdg{} retrieve the planted coupling on all six datasets, since \milag{} is non-parametric and also evaluated at lag $0$, and \cdg{} makes no assumption on the functional form and sees all channels jointly. We therefore use them to profile the real-world datasets.

\begin{table*}[t]
\centering
\newcommand{\miss}{\color{red!80!black}}%
\newcommand{\ds}[1]{\textsf{#1}}%
\renewcommand{\arraystretch}{1.15}
\caption{Retrieval details behind Table~\ref{tab:synth_retrieval}, one row per planted source. \cdg{}: gain of TSMixer over TSMixer\textsubscript{CI} on $c_0$ minus that on the \ctype{independent} reference (pp), over forecast steps $h\leq\ell$ ($\ell{=}24$ if no lag is planted). Mean$\pm$std over 3 seeds, all significant (one-sided Welch $t$-test, $p<0.05$). $\hat{\ell}$ and $r$ as in Eq.~\eqref{eq:synth-retrieval}, with a source flagged if $r>1$. GC gives no lag. ``${<}\,$null'': not flagged ($r\leq 1$). Red: missed, mis-located or false.}
\label{tab:synth_retrieval_detailed}
\footnotesize
\begin{tabular}{l c r@{${}\pm{}$}l cc cc c}
\toprule
 & & \multicolumn{2}{c}{} & \multicolumn{2}{c}{\milag{}} & \multicolumn{2}{c}{TE} & GC \\
\cmidrule(lr){5-6}\cmidrule(lr){7-8}\cmidrule(lr){9-9}
Dataset & Source & \multicolumn{2}{c}{\cdg{} (pp)} & $\hat{\ell}$ & $r$ & $\hat{\ell}$ & $r$ & $r$ \\
\midrule
\ds{independent} & none & \multicolumn{2}{c}{reference} & \multicolumn{5}{c}{no source flagged} \\
\ds{contemp-linear} & $c_1$ & $+1.5$ & $0.5$ & 0 & 14 & \miss 24 & \miss 11 & 1.5 \\
\ds{lagged-linear} & $c_1$ & $+25.8$ & $0.4$ & 24 & 13 & 24 & 74 & 51 \\
\ds{lagged-non-linear} & $c_1$ & $+4.8$ & $0.6$ & 12 & 17 & 12 & 14 & \miss ${<}\,$null \\
\ds{multiplicative-joint} & $c_1$ & $+5.2$ & $2.5$ & 6 & 3.2 & 6 & 4.3 & \miss 1.1 \\
 & $c_2$ & \multicolumn{2}{c}{} & 6 & 3.5 & 6 & 4.1 & \miss ${<}\,$null \\
\ds{additive-mixed} & $c_1$ & $+24.1$ & $0.2$ & 24 & 16 & 24 & 73 & 41 \\
 & $c_2$ & \multicolumn{2}{c}{} & 8 & 9.2 & 8 & 7.9 & \miss ${<}\,$null \\
\bottomrule
\end{tabular}
\end{table*}

\section{Channel Dependency Measurement of Standard Datasets}
\label{sec:std-measurement}

We apply \milag{} and \cdg{} to the standard datasets, namely ETTh1, ETTh2, ETTm1, ETTm2 \citep{zhou2021informer}, Electricity, Traffic, ExchangeRate, SolarAL \citep{lai2018modeling}, Weather, and ILI \citep{wu2021autoformer}. As a reference, and similar to \citet{abdelmalak2026channel}, we also include six datasets from the chaotic ODE benchmark of \citet{gilpin2021chaos}, whose coupling is determined by the underlying ODEs (Appendix~\ref{sec:ode-datasets-details}). Table~\ref{tab:analysis_standard} reports both measures.

\begin{table}[t]
\centering
\small
\caption{Channel dependency of the standard datasets and of the chaotic ODE benchmark. \milag{}: share of channel pairs coupled more strongly at a lag than at lag 0. \cdg{}: test-MSE reduction of TSMixer over TSMixer\textsubscript{CI} (Eq.~\ref{eq:cd-gain}), with MSE averaged over horizons and 3 seeds (Table~\ref{tab:seed_mse_mean}, ODE MSEs from \citet{abdelmalak2026channel}). \textbf{Bold}: positive \cdg{} and the better TSMixer variant.}
\label{tab:analysis_standard}
\begin{tabular}{l cc cc}
\toprule
Dataset & \milag{} (\%) & \cdg{} (\%) & TSMixer & TSMixer\textsubscript{CI} \\
\midrule
\multicolumn{5}{l}{\textit{Standard MTSF}} \\
\quad ETTh1 & 19 & $-14.9$ & 0.498 & \textbf{0.433} \\
\quad ETTh2 & 19 & $-5.4$ & 0.411 & \textbf{0.390} \\
\quad ETTm1 & 38 & $-4.8$ & 0.369 & \textbf{0.352} \\
\quad ETTm2 & 38 & $-2.0$ & 0.271 & \textbf{0.266} \\
\quad Weather & 52 & $-1.0$ & 0.235 & \textbf{0.232} \\
\quad Electricity & 61 & $-5.1$ & 0.166 & \textbf{0.158} \\
\quad Traffic & 27 & $-9.2$ & 0.427 & \textbf{0.391} \\
\quad ExchangeRate & 7 & $-1.3$ & 0.154 & \textbf{0.152} \\
\quad ILI & 17 & $-6.3$ & 1.992 & \textbf{1.873} \\
\quad SolarAL & 14 & $-4.0$ & 0.200 & \textbf{0.193} \\
\cmidrule(l){2-3}
\quad \textit{category median} & \textit{23} & $\mathit{-4.9}$ & & \\
\midrule
\multicolumn{5}{l}{\textit{Chaotic ODE (known-correlated reference)}} \\
\quad Lorenz & 83 & $\mathbf{+1.4}$ & \textbf{0.868} & 0.880 \\
\quad Lorenz-Coupled & 63 & $\mathbf{+5.3}$ & \textbf{0.900} & 0.950 \\
\quad Blinking-Rotlet & 83 & $\mathbf{+16.0}$ & \textbf{0.487} & 0.580 \\
\quad CellCycle & 67 & $\mathbf{+2.7}$ & \textbf{0.771} & 0.792 \\
\quad Double-Pendulum & 92 & $\mathbf{+4.0}$ & \textbf{0.737} & 0.768 \\
\quad Hopfield & 73 & $\mathbf{+14.2}$ & \textbf{0.435} & 0.507 \\
\cmidrule(l){2-3}
\quad \textit{category median} & \textit{78} & $\mathit{+4.6}$ & & \\
\bottomrule
\end{tabular}
\end{table}

The two measures clearly separate the two groups. On \milag{}, the standard datasets have a median of only 23\% lagged-coupled channel pairs, and only Weather (52\%) and Electricity (61\%) score above 50\%. The ODE datasets reach a median of 78\%, with none below 63\%. \cdg{} confirms this picture from the model-based measurement perspective. TSMixer\textsubscript{CI} outperforms TSMixer on all 10 standard datasets (median \cdg{} of $-4.9\%$), while the ODE datasets reach a median of $+4.6\%$ and up to $+16.0\%$ (Blinking-Rotlet). We therefore conclude that the current standard MTSF datasets lack exploitable cross-channel coupling. Discarding the other channels does not hurt forecasting on them, and even consistently improves it. This raises the question of whether we can find real-world datasets with richer cross-channel coupling, where multivariate models can actually benefit from channel mixing.

\begin{table*}[t]
\centering
\caption{The coupled datasets. \emph{Timesteps}: series length, pooled over the independent series for the multi-series panels noted in \emph{Domain}. \emph{Granularity}: sampling interval.}
\label{tab:datasets}
\setlength{\tabcolsep}{6pt}
\begin{tabular}{@{}lrrll@{}}
\toprule
\textbf{Dataset} & \textbf{Channels} & \textbf{Timesteps} & \textbf{Granularity} & \textbf{Domain} \\
\midrule
AirQuality          & 132 & 35{,}064  & 1 hour  & Environment \\
AQShunyi            & 11  & 35{,}064  & 1 hour  & Environment \\
CzeLan              & 11  & 19{,}934  & 30 min  & Environment \\
ZafNoo              & 11  & 19{,}225  & 30 min  & Environment \\
Bavaria-Iller       & 4   & 59{,}502  & 15 min  & Hydrology \\
Bavaria-Donau       & 11  & 31{,}250  & 15 min  & Hydrology \\
WaterQuality-Darwin & 6   & 106{,}606 & 15 min  & Marine (7 sites) \\
Current-Velocity    & 6   & 174{,}425 & 20 min  & Ocean (27 deployments) \\
Wearable            & 24  & 38{,}913  & 5 min   & Health \\
HouseholdPower      & 7   & 138{,}352 & 15 min  & Energy \\
\bottomrule
\end{tabular}
\end{table*}

\section{The MixBench-TS Benchmark}
\label{sec:new-benchmark}

\subsection{Datasets}
\label{sec:new-datasets}
We propose MixBench-TS, a benchmark of 10 real-world datasets collected from recent MTSF benchmarks and public data repositories. We refer to these datasets as \emph{the coupled datasets}. AQShunyi, CzeLan, and ZafNoo are taken from TFB \citep{qiu2024tfb}, WaterQuality-Darwin and Current-Velocity from TIME \citep{qiao2026its}, and Bavaria-Iller and Bavaria-Donau from CausalRivers \citep{stein2025causalrivers}. Wearable is taken from \citet{baigutanova2025continuous}, AirQuality from the Beijing multi-site air-quality data \citep{zhang2017cautionary}, and HouseholdPower from the UCI household power consumption data \citep{hebrail2006individual}. Although publicly available, these datasets are rarely used for MTSF evaluation. Since our aim is to complement the standard datasets with datasets that have exploitable cross-channel coupling, we screen candidate datasets with \cdg{} (Section~\ref{sec:measurement-evaluation}) and retain only those on which channel mixing pays off ($\cdg{}>0$). \milag{} is not used for this selection and only characterizes the type of coupling. They cover a wide range of domains (environmental monitoring, energy, hydrology, marine sensing, and health), numbers of channels (4 to 132), series lengths (19{,}225 to 174{,}425 timesteps), and sampling intervals (5 min to 1 hour), as summarized in Table~\ref{tab:datasets}. The seven datasets on which lagged coupling also dominates ($\milag{}>50\%$) are exactly those with the highest \cdg{} ($\geq +1.1\%$). We call them \emph{MixBench-TS-Core}, a stricter subset that we consider the strongest testbed for multivariate models. We additionally report two spatio-temporal traffic datasets (METR-LA and PEMS-BAY) \citep{li2018diffusion} in Appendix~\ref{sec:spatiotemporal-datasets}.

\subsection{Experimental Protocol}
\label{sec:evaluation-metrics-hp-tuning}
We compare six state-of-the-art models covering both channel strategies (Section~\ref{sec:related-work}): DLinear \citep{zeng2023transformers}, TSMixer\textsubscript{CI} \citep{chen2023tsmixer}, and CycleNet \citep{lin2024cyclenet} as CI models, and TSMixer \citep{chen2023tsmixer}, iTransformer \citep{liu2024itransformer}, and SimpleTM \citep{chen2025simpletm} as CD models. Since insufficient tuning can distort model comparisons \citep{brigato2025there}, we tune every model on every dataset with Optuna's TPE sampler \citep{akiba2019optuna,bergstra2011algorithms} for 20 trials. We use the search space of the original paper or, if unavailable, a range around its default parameters (Appendix~\ref{sec:hyperparameter-tuning-setup}). Following \citet{abdelmalak2026channel}, we additionally tune the lookback window. Candidate lookback windows and forecast horizons are chosen per dataset based on its sampling interval (Appendix~\ref{sec:datasets-details}). All models are trained with the MSE loss, selected by validation MSE, and evaluated by test MSE and MAE, averaged over horizons and three seeds.

\begin{table*}[t]
\centering\small
\setlength{\tabcolsep}{3.6pt}
\caption{Channel dependency and test MSE on the coupled datasets, sorted by \cdg{}. \milag{} and \cdg{} as in Table~\ref{tab:analysis_standard}. MSE averaged over horizons and 3 seeds, best per row \textbf{bold}, second \underline{underlined}. Win: channel strategy of the best model. \colorbox{ciband}{\catCI}: TSM\textsubscript{CI}\,=\,TSMixer\textsubscript{CI}, DLin.\,=\,DLinear, CycNet\,=\,CycleNet. \colorbox{cdband}{\catCD}: TSM\,=\,TSMixer, SimTM\,=\,SimpleTM, iTr.\,=\,iTransformer. Wins: number of best rows (ties counted for each). Above the line: MixBench-TS-Core ($\milag{}>50\%$). All datasets in Table~\ref{tab:seed_mse_mean}.}
\label{tab:occasional_results}
\begin{tabular}{l cc >{\columncolor{ciband}}c >{\columncolor{ciband}}c >{\columncolor{ciband}}c >{\columncolor{cdband}}c >{\columncolor{cdband}}c >{\columncolor{cdband}}c c}
\toprule
  & & & \multicolumn{3}{c}{\cellcolor{ciband}\textbf{CI}} & \multicolumn{3}{c}{\cellcolor{cdband}\textbf{CD}} &   \\
\cmidrule(lr){4-6}\cmidrule(lr){7-9}
Dataset & \milag{} (\%) & \cdg{} (\%) & TSM\textsubscript{CI} & DLin. & CycNet & TSM & SimTM & iTr. & Win \\
\midrule
Bavaria-Donau & 78 & $+11.1$ & 0.097 & \underline{0.095} & 0.098 & \textbf{0.086} & \textbf{0.086} & 0.100 & \catCD \\
Bavaria-Iller & 67 & $+3.4$ & 0.185 & \underline{0.176} & 0.186 & 0.179 & \textbf{0.170} & 0.186 & \catCD \\
Current-Velocity & 83 & $+3.1$ & 0.263 & 0.264 & 0.270 & \underline{0.255} & \textbf{0.254} & 0.265 & \catCD \\
CzeLan & 82 & $+3.0$ & 0.212 & 0.218 & 0.216 & \textbf{0.205} & \underline{0.211} & 0.225 & \catCD \\
ZafNoo & 57 & $+1.8$ & 0.523 & \textbf{0.486} & 0.528 & \underline{0.514} & 0.529 & 0.539 & \catCI \\
AirQuality & 54 & $+1.5$ & 0.750 & \textbf{0.738} & 0.744 & \underline{0.739} & 0.752 & 0.748 & \catCI \\
AQShunyi & 53 & $+1.1$ & 0.684 & \underline{0.672} & 0.682 & 0.677 & \textbf{0.664} & 0.697 & \catCD \\
\midrule
WaterQuality-Darwin & 40 & $+0.8$ & \underline{0.270} & 0.281 & 0.271 & \textbf{0.268} & 0.275 & 0.280 & \catCD \\
Wearable & 17 & $+0.7$ & 0.671 & \underline{0.668} & 0.781 & \textbf{0.666} & 0.753 & 0.755 & \catCD \\
HouseholdPower & 17 & $+0.6$ & 0.652 & \textbf{0.644} & \underline{0.646} & 0.648 & 0.648 & \underline{0.646} & \catCI \\
\midrule
\textit{median / wins} & \textit{55.5} & $\mathit{+1.7}$ & 0 & \underline{3} & 0 & \textbf{4} & \textbf{4} & 0 & \\
\bottomrule
\end{tabular}
\end{table*}

\subsection{Results}
\label{sec:new-benchmark-results}
\label{sec:MTSF-experiments-results}

Table~\ref{tab:occasional_results} combines the two dependency measures with the forecasting results on the coupled datasets. By construction, channel mixing helps on all of them (median \cdg{} of $+1.7\%$ vs.\ $-4.9\%$ on the standard datasets), and they have a median of 55.5\% lagged-coupled pairs, compared to 23\%. The winning model class follows the two measures. The four datasets with the highest \cdg{} (Bavaria-Donau, Bavaria-Iller, Current-Velocity, and CzeLan) all combine a high share of lagged coupling (67--83\%) with a CD model winning. Overall, CI models win only 3 of the 10 coupled datasets by MSE and 2 of 10 by MAE, compared to 10 of 10 and 8 of 10 standard datasets (Tables~\ref{tab:seed_mse_mean} and~\ref{tab:seed_mae_mean}).

There are, however, a few outliers. On ZafNoo, AirQuality, and HouseholdPower, channel mixing improves TSMixer ($+0.6$ to $+1.8\%$), yet DLinear achieves the lowest MSE, suggesting that the benefit of channel mixing is smaller than the differences between architectures there. Conversely, TSMixer wins on Wearable and WaterQuality-Darwin, although both fall outside MixBench-TS-Core. Finally, DLinear remains a strong baseline, which indicates that more complex architectures do not yet fully exploit these datasets and further motivates including them in MTSF benchmarks.

\section{Conclusions and Future Work}
\label{sec:conclusion}
In this work, we asked two questions. First, how can we reliably measure lagged, non-linear, and joint coupling in MTSF datasets? On synthetic datasets with known ground-truth coupling, only lagged MI (\milag{}) and CD gain (\cdg{}) retrieve all planted couplings. Second, do the standard MTSF datasets have such coupling? The answer is no. They have a median of only 23\% lagged-coupled channel pairs compared to 78\% in the chaotic ODE benchmark, and channel mixing hurts on every one of them. We therefore proposed MixBench-TS, a benchmark of 10 real-world coupled datasets for evaluating multivariate forecasting models. These datasets have a median of 55.5\% lagged-coupled pairs and a positive \cdg{}, and CI models win on only 3 of them, compared to all 10 standard datasets. We recommend profiling further datasets with \milag{} and \cdg{}. Those with $\cdg{}>0$ can join MixBench-TS, and those that also have $\milag{}>50\%$ can join MixBench-TS-Core. Evaluating multivariate models requires benchmarks that actually contain multivariate structure, so that the benefits of channel mixing can be observed rather than assumed.

\paragraph{Limitations.}
\label{sec:limitations}
First, \cdg{} uses a single model family (TSMixer) and thus also reflects whether this architecture can exploit the coupling, e.g., Electricity has 61\% lagged pairs but a \cdg{} of $-5.1\%$. Second, \milag{} is a pairwise fraction that ignores coupling strength, captures joint coupling only indirectly, and uses at most 40 channels on wide datasets. Third, our synthetic suite has only four channels and fixed couplings, without the regime changes and hidden common drivers of real data.

\paragraph{Future Work.}
The landmarking concept behind \cdg{} can be extended beyond TSMixer to further CD models and their CI variants, making \cdg{} less dependent on a single architecture. Per-channel and time-windowed coupling maps could guide selective channel mixing and capture couplings that change across regimes. Finally, our criteria can broaden MixBench-TS with short-and-wide, spatio-temporal, and irregularly sampled datasets.

\section*{AI Use Statement}
We used generative AI tools for three purposes. First, we used them to polish the writing. Second, we used them to draft parts of individual sections, which we then revised over several rounds of proofreading. Third, we used them to help search for related work. We verified every retrieved reference and took care that no closely related work is omitted. We reviewed and validated all content of the final version and take full responsibility for it.

\section*{Reproducibility Statement}
We describe all components needed to reproduce our results. The datasets, their sources, splits, horizons, and lookback windows are given in Section~\ref{sec:new-datasets} and Appendices~\ref{sec:datasets-details} and~\ref{sec:problem-formulation-detailed}. The coupling measures are defined in Section~\ref{sec:statistical-profiling}, with implementation details in Appendix~\ref{sec:stats-profiling-implementation-details}. The synthetic data generation, including all generation parameters, is described in Appendix~\ref{sec:synthetic-data-generation-details}. The models, the tuning protocol, and all hyperparameter search ranges are given in Section~\ref{sec:evaluation-metrics-hp-tuning} and Appendix~\ref{sec:hyperparameter-tuning-setup}. All forecasting results are averaged over three seeds (2021, 2025, and 2026), with standard deviations in Appendix~\ref{sec:random-seed-full-results}. Our anonymized repository (\url{https://anonymous.4open.science/r/mixbench-ts-B027}) contains the code for all experiments, the profiling toolkit, and the synthetic data generator, together with instructions for obtaining each dataset and running the experiments.

\bibliography{main}
\bibliographystyle{iclr2027_conference}

\appendix

\section{Datasets - More Details}
\label{sec:datasets-details}

All forecasting horizons $H$ and lookback windows $L$ are given in timesteps of each
dataset's sampling interval, and the lookback is tuned jointly with the other
hyperparameters (Appendix~\ref{sec:hyperparameter-tuning-setup}).

\paragraph{Data splits.}
For the ETT datasets, we follow the standard protocol of 12/4/4 months for training, validation, and testing. All other single-series datasets are split chronologically into 70/10/20\%. For the multi-series datasets (WaterQuality-Darwin and Current-Velocity), each series is split 70/10/20\% in time, and the windows of all series are pooled. Wearable is split at the subject level (70/15/15\%, i.e., 34/7/8 of 49 subjects, assigned at random with a fixed seed), and no window spans two subjects. Validation and test windows may use lookback values from the preceding split, but their forecast targets always lie within their own split. Channels are standardized with training-set statistics, and all metrics are reported on the standardized scale.

\paragraph{Standard real-world datasets.}
For comparability with prior work, we follow the established long-term
forecasting protocol for these datasets: $H\in\{96,192,336,720\}$ with
$L\in\{48,96,192,336,720\}$, except for ExchangeRate ($H\in\{30,90,180,365\}$,
$L\in\{30,60,90,180,365\}$) and ILI ($H\in\{24,36,48\}$,
$L\in\{12,24,36,52,104\}$).
\begin{itemize}
\item \textbf{ETTh1, ETTh2, ETTm1, ETTm2} (7 channels; hourly and 15-minute).
  Six power-load features (high, middle and low useful and useless load) and the
  oil temperature of two electricity transformers in China (2016--2018).
\item \textbf{Weather} (21 channels; 10-minute). Meteorological indicators
  (pressure, temperatures, humidity, wind, precipitation, radiation) of a
  weather station in Jena, Germany.
\item \textbf{Electricity} (321 channels; hourly). Electricity consumption of 321
  clients (2012--2014).
\item \textbf{Traffic} (862 channels; hourly). Road occupancy rates of 862 freeway
  sensors in the San Francisco Bay Area (2015--2016).
\item \textbf{ExchangeRate} (8 channels; daily). Daily exchange rates of eight
  currencies (1990--2016).
\item \textbf{ILI} (7 channels; weekly). Influenza-like-illness surveillance
  indicators of the US CDC (weighted and unweighted ILI rates, patient counts by
  age group, number of providers).
\item \textbf{SolarAL} (137 channels; 10-minute). Solar power output of 137
  photovoltaic plants in Alabama (2006).
\end{itemize}

\paragraph{Spatio-temporal (traffic) datasets.}
\begin{itemize}
\item \textbf{METR-LA} (207 channels) and \textbf{PEMS-BAY} (325 channels; both
  5-minute). Traffic speed from loop detectors on the highways of Los Angeles
  (2012) and the San Francisco Bay Area (2017). $L\in\{48,96,192,288\}$ (4\,h to 1
  day), $H\in\{12,48,144,288\}$, i.e.\ 1\,h (the standard short-term traffic
  horizon), 4\,h, 12\,h and one full day, which spans the morning and evening peaks.
\end{itemize}

\paragraph{Coupled real-world datasets.}
Five of these datasets are part of recent forecasting benchmarks: AQShunyi,
CzeLan and ZafNoo of TFB \citep{qiu2024tfb}, whose horizons
$H\in\{96,192,336,720\}$ we adopt, and WaterQuality-Darwin and Current-Velocity of
TIME \citep{qiao2026its}, which targets zero-shot evaluation of foundation models.
For the remaining horizons and all lookback windows, the choices cover each
dataset's dominant natural cycles, from within-cycle to multi-cycle forecasts, so
that each horizon corresponds to an operationally meaningful lead time.
\begin{itemize}
\item \textbf{AirQuality} (132 channels; hourly). Eleven air-quality and
  meteorological variables (PM2.5, PM10, SO$_2$, NO$_2$, CO, O$_3$, temperature,
  pressure, dew point, rain, wind speed) at each of 12 monitoring stations in
  Beijing (2013--2017). The stations share pollution episodes and weather.
  $L\in\{48,\dots,720\}$, $H\in\{96,192,336,720\}$ (4 to 30 days), covering the
  multi-day build-up and dispersal of pollution episodes.
\item \textbf{AQShunyi} (11 channels; hourly). The same eleven variables at the
  single Shunyi station, where pollutants are coupled with each other and with the
  local meteorology. $L\in\{24,48,168,336,720\}$ (1 day to 30 days), $H$ as for
  AirQuality.
\item \textbf{CzeLan} and \textbf{ZafNoo} (11 channels each; 30-minute). The sap
  flow of a single tree together with its environmental drivers (incoming
  short-wave, extraterrestrial and photosynthetically active radiation, air
  temperature, relative humidity, vapor-pressure deficit, precipitation, wind
  speed, shallow and deep soil water content), at a site in the Czech Republic
  and one in South Africa. Transpiration responds to radiation and
  vapor-pressure deficit with a lag, and to soil moisture on longer scales.
  $L\in\{48,96,336,672\}$ (1 day to 2 weeks), $H\in\{96,192,336,720\}$ (2 to 15
  days), i.e.\ from a few diurnal cycles to about two weeks of drying or rewetting.
\item \textbf{Bavaria-Iller} (4 channels) and \textbf{Bavaria-Donau} (11 channels;
  both 15-minute). River discharge at gauging stations along the Iller and
  Danube river networks in Bavaria, where flow at a gauge propagates to the
  gauges further along the network with a travel-time delay.
  $L\in\{48,96,192,336,720\}$ (12\,h to 7.5 days), $H\in\{24,96,336\}$, i.e.\ 6\,h,
  1 day and 3.5 days: the lead times of short-term flood warning, next-day
  operation and multi-day flood-wave routing.
\item \textbf{WaterQuality-Darwin} (6 channels; 15-minute; 7 series). Coastal
  water quality at the Darwin national reference station, Australia: electrical
  conductivity, dissolved oxygen, salinity, temperature, turbidity and
  chlorophyll, which are coupled through tidal mixing and biological activity.
  $L\in\{48,96,192,336\}$ (12\,h to 3.5 days), $H\in\{24,96\}$, i.e.\ 6\,h (about
  half a semi-diurnal tidal cycle) and 1 day (two tidal and one diurnal cycle).
\item \textbf{Current-Velocity} (6 channels; 20-minute; 27 series). ADCP
  mooring records from the Australian national mooring network: zonal,
  meridional and vertical current velocity, water temperature, relative pressure
  and acoustic backscatter intensity. Currents and pressure are coupled through
  the tides. $L\in\{48,96,192,336\}$ (16\,h to 4.7 days), $H\in\{24,96\}$, i.e.\ 8\,h
  (within one semi-diurnal tidal cycle) and 32\,h (more than two tidal cycles).
\item \textbf{Wearable} (24 channels; 5-minute). Wrist-worn sensor features of
  free-living participants: heart rate, inter-beat interval, heart-rate-variability
  statistics (SDNN, SDSD, RMSSD, pNN20, pNN50, LF, HF, LF/HF), accelerometer,
  gravity and gyroscope averages, steps, distance, calories and ambient light.
  Physical activity drives the cardiac channels. $L\in\{48,96,192,288,576\}$ (4\,h
  to 2 days), $H\in\{12,48,144,288\}$, i.e.\ 1\,h, 4\,h, 12\,h and one day: from the
  next activity bout to the full circadian cycle.
\item \textbf{HouseholdPower} (7 channels; 15-minute). Electricity consumption of
  a single household (2006--2010): global active and reactive power, voltage,
  current intensity and three sub-meterings (kitchen, laundry, water heater and
  air conditioning), where the total load is the sum of its components.
  $L\in\{48,96,192,336,672\}$ (12\,h to 1 week), $H\in\{96,192,336,672\}$, i.e.\ 1,
  2, 3.5 and 7 days: from the daily usage cycle to the weekly one.
\end{itemize}

\paragraph{Chaotic ODE systems.}
\label{sec:ode-datasets-details}
The six systems are taken from the \texttt{dysts} forecasting benchmark of
\citet{gilpin2021chaos} and were used as a known-correlated reference for MTSF by
\citet{abdelmalak2026channel}, whose setup we follow and whose TSMixer and TSMixer\textsubscript{CI} test MSEs we report in Table~\ref{tab:analysis_standard}. Each system is simulated over
$t\in[0,3000]$ with 60{,}000 samples ($\Delta t\approx0.05$), and its channels are
the state variables of the system (Table~\ref{tab:datasets_ode}). Since the
coupling is given by the governing equations, every channel depends on the others
by construction. $L\in\{48,96,192,336,720\}$ and $H\in\{96,192,336,720\}$, as for
the standard datasets.
\begin{itemize}
\item \textbf{Lorenz} (3 channels). A minimal weather model based on atmospheric convection (Lorenz-63).
\item \textbf{Lorenz-Coupled} (6 channels). Two coupled Lorenz oscillators.
\item \textbf{Blinking-Rotlet} (3 channels). Chaotic advection by a blinking
  rotlet mixer, a model of fluid mixing, in polar coordinates.
\item \textbf{Double-Pendulum} (4 channels). Two coupled rigid pendula without
  damping.
\item \textbf{CellCycle} (6 channels). A simplified model of the cell cycle.
\item \textbf{Hopfield} (6 channels). A Hopfield neural network with frustrated
  connectivity.
\end{itemize}

\begin{table}[h]
\centering
\caption{The chaotic ODE systems, used as a known-correlated reference in Table~\ref{tab:analysis_standard}. Columns as in Table~\ref{tab:datasets}, with \emph{Granularity} the integration step $\Delta t$ and \emph{Domain} the modeled system.}
\label{tab:datasets_ode}
\setlength{\tabcolsep}{6pt}
\begin{tabular}{@{}lrrll@{}}
\toprule
\textbf{System} & \textbf{Channels} & \textbf{Timesteps} & \textbf{Granularity} & \textbf{Domain} \\
\midrule
Lorenz          & 3 & 60{,}000 & $\Delta t{=}0.05$ & Chaos \\
Lorenz-Coupled  & 6 & 60{,}000 & $\Delta t{=}0.05$ & Coupled chaos \\
Blinking-Rotlet & 3 & 60{,}000 & $\Delta t{=}0.05$ & Fluid mixing \\
Double-Pendulum & 4 & 60{,}000 & $\Delta t{=}0.05$ & Mechanics \\
CellCycle       & 6 & 60{,}000 & $\Delta t{=}0.05$ & Biology \\
Hopfield        & 6 & 60{,}000 & $\Delta t{=}0.05$ & Neural \\
\bottomrule
\end{tabular}
\end{table}

\section{Spatio-Temporal Datasets}
\label{sec:spatiotemporal-datasets}

In addition to the standard and coupled datasets, we evaluate two spatio-temporal traffic datasets \citep{li2018diffusion}: METR-LA (207 channels, 34{,}272 timesteps) and PEMS-BAY (325 channels, 52{,}116 timesteps), both sampled every 5 minutes. Since their channels correspond to road sensors of the same traffic network, cross-channel (spatial) coupling is expected. However, they are neither part of the standard datasets nor of our proposed coupled datasets, which is why we report them separately. Table~\ref{tab:st_analysis} shows that both datasets together have a similar median share of lagged coupling as the coupled datasets (53.5\% vs.\ 55.5\%), well above the standard datasets (23\%), but a mixed \cdg{}: channel mixing helps on PEMS-BAY ($+6.4\%$) but not on METR-LA ($-1.5\%$). The forecasting results in Table~\ref{tab:seed_mse_mean} follow the same pattern, with a CD model winning on PEMS-BAY and a CI model winning on METR-LA in terms of MSE. The corresponding MAE results are given in Table~\ref{tab:seed_mae_mean} and the per-horizon results in Table~\ref{tab:perh_spatiotemporal}.

\begin{table}[h]
\centering
\small
\caption{Channel dependency of the spatio-temporal datasets. Columns as in Table~\ref{tab:analysis_standard}.}
\label{tab:st_analysis}
\begin{tabular}{l cc cc}
\toprule
Dataset & \milag{} (\%) & \cdg{} (\%) & TSMixer & TSMixer\textsubscript{CI} \\
\midrule
\quad METR-LA & 38 & $-1.5$ & 0.930 & \textbf{0.916} \\
\quad PEMS-BAY & 69 & $\mathbf{+6.4}$ & \textbf{0.469} & 0.501 \\
\cmidrule(l){2-3}
\quad \textit{category median} & \textit{53.5} & $\mathit{+2.4}$ & & \\
\bottomrule
\end{tabular}
\end{table}

\section{Problem Formulation - Detailed}
\label{sec:problem-formulation-detailed}

Each dataset is a multivariate time series
$\mathbf{Z} = (\mathbf{z}_1, \dots, \mathbf{z}_T) \in \mathbb{R}^{T \times C}$
observed at $T$ equally spaced steps, where $C$ is the number of channels
(variates) and $\mathbf{z}_t \in \mathbb{R}^{C}$ represents all channel values
at step $t$. $\mathbf{Z}$ is split \emph{chronologically} into three contiguous
segments: training $\mathbf{Z}_{\mathrm{tr}} = (\mathbf{z}_1, \dots, \mathbf{z}_{T_{\mathrm{tr}}})$,
validation $\mathbf{Z}_{\mathrm{val}} = (\mathbf{z}_{T_{\mathrm{tr}}+1}, \dots, \mathbf{z}_{T_{\mathrm{val}}})$,
and test $\mathbf{Z}_{\mathrm{te}} = (\mathbf{z}_{T_{\mathrm{val}}+1}, \dots, \mathbf{z}_T)$. The
boundaries $T_{\mathrm{tr}} = \lfloor \rho_{\mathrm{tr}} T \rfloor$ and
$T_{\mathrm{val}} = \lfloor (\rho_{\mathrm{tr}} + \rho_{\mathrm{val}}) T \rfloor$
are set by the split fractions
$\rho_{\mathrm{tr}} + \rho_{\mathrm{val}} + \rho_{\mathrm{te}} = 1$, so that
every training step precedes every validation step, which in turn precedes
every test step. We write $\mathcal{T}_{\mathrm{tr}} = \{L, \dots, T_{\mathrm{tr}}-H\}$,
$\mathcal{T}_{\mathrm{val}} = \{T_{\mathrm{tr}}, \dots, T_{\mathrm{val}}-H\}$, and
$\mathcal{T}_{\mathrm{te}} = \{T_{\mathrm{val}}, \dots, T-H\}$ for the corresponding
sets of time anchors, which represent the indices of the different time series samples where $L$ and $H$ are the \emph{lookback window} and \emph{forecast horizon} lengths, respectively.

Supervised samples are constructed by sliding a
fixed-length window over the series (we always use a stride of 1 throughout our experiments).
A sample anchored at step $t$ splits into a lookback window and a forecast horizon as follows,
\begin{equation}
  \mathbf{X}^{(t)} = (\mathbf{z}_{t-L+1}, \dots, \mathbf{z}_{t}) \in \mathbb{R}^{L \times C},
  \qquad
  \mathbf{Y}^{(t)} = (\mathbf{z}_{t+1}, \dots, \mathbf{z}_{t+H}) \in \mathbb{R}^{H \times C}.
\end{equation}
We index each sample by its \emph{anchor} $t$, the last observed step, so that the
lookback spans steps $t-L+1, \dots, t$ and the forecast spans $t+1, \dots, t+H$.
Each sample is assigned to the split whose anchor set $\mathcal{T}_{\mathrm{split}}$
contains it, giving
\begin{equation}
  \mathcal{D}_{\mathrm{split}} = \{\, (\mathbf{X}^{(t)}, \mathbf{Y}^{(t)}) : t \in \mathcal{T}_{\mathrm{split}} \,\},
  \qquad \mathrm{split} \in \{\mathrm{tr}, \mathrm{val}, \mathrm{te}\},
\end{equation}
with the anchor ranges chosen so that every lookback fits within the series and
every forecast horizon stays inside its split. Consequently, forecast
targets never cross a split boundary, so no future information leaks into training
or evaluation, even though a lookback window may reach back into the preceding
segment, as is standard in the forecasting literature. A model
$f_{\boldsymbol{\theta}} : \mathbb{R}^{L \times C} \to \mathbb{R}^{H \times C}$
predicts
$\widehat{\mathbf{Y}}^{(t)} = f_{\boldsymbol{\theta}}(\mathbf{X}^{(t)})$
jointly over all $C$ channels, and its parameters are fit by minimizing the
empirical training loss
$\boldsymbol{\theta}^{\star} = \arg\min_{\boldsymbol{\theta}} \frac{1}{|\mathcal{D}_{\mathrm{tr}}|} \sum_{(\mathbf{X}^{(t)}, \mathbf{Y}^{(t)}) \in \mathcal{D}_{\mathrm{tr}}} \ell(\widehat{\mathbf{Y}}^{(t)}, \mathbf{Y}^{(t)})$
for a per-sample loss $\ell$ (e.g.\ mean squared error). The MTSF benchmarks commonly use a horizon
$H \in \{96, 192, 336, 720\}$. However, as we show in Appendix~\ref{sec:datasets-details}, we adapt
these horizons to be of semantic relevance for the remaining datasets.

\section{Statistical Profiling - Implementation Details}
\label{sec:stats-profiling-implementation-details}

In this appendix, we provide more detailed implementation details which should be helpful in both reproducing the results and understanding our design choices for these metrics. For datasets with more than $40$ channels, all three pairwise measures are computed on a subset of $40$ channels drawn by uniform random sampling without replacement. A sensitivity analysis over this subsampling is reported in Appendix~\ref{sec:parameter-sensitivity}.

\subsection{Granger Causality}

Granger Causality was originally proposed by \citet{granger1969investigating} and has been widely used in econometrics and time series analysis. The basic idea is that if a time series $X$ Granger-causes another time series $Y$, then past values of $X$ should contain information that helps predict $Y$ beyond the information contained in past values of $Y$ alone. What makes this method more challenging to apply in practice are the following factors:

\begin{itemize}
  \item \textbf{Stationarity}: Granger Causality assumes that the time series are stationary. In practice, many real-world time series exhibit trends and seasonality, which can violate this assumption. To address this, we apply seasonal differencing and regular differencing to make the series stationary before applying Granger Causality tests.
  \item \textbf{Lag Selection}: The choice of lag length is crucial in Granger Causality analysis. For comparability with deep learning models, we set the VAR order $p$ equal to the smallest candidate lookback window of the dataset (the order $p$ of the main text). This allows us to capture the relevant temporal dependencies without introducing unnecessary complexity.
\end{itemize}

The seasonal period $P$ is detected once per dataset, before any channel subsampling, as the strongest local peak of the autocorrelation function averaged over all detrended channels. It therefore does not depend on the channel subset, and both series of a pair receive the same seasonal difference at period $P$. For the multi-series panels, we use a fixed period instead ($P=96$ for WaterQuality-Darwin and $P=72$ for Current-Velocity), and the synthetic lag profiles use $P=24$. The seasonal difference is skipped if the series is shorter than $3P$. We then use the Augmented Dickey-Fuller (ADF) \citep{said1984testing} test to determine if additional regular differencing is needed. The ADF test checks for the presence of a unit root in the time series, which indicates non-stationarity. For each pair, the ADF test selects a differencing order $d\leq 2$ separately for the source and the target, and both series are then differenced $\max(d_{\mathrm{src}}, d_{\mathrm{tgt}})$ times. This common order keeps the two series aligned and equally transformed.

With these settings fixed, both models are fitted with Ordinary Least Squares (OLS) on the preprocessed series. The sums of squared residuals ($\mathrm{SSR}$) of the restricted and multivariate models are then used to compute the degrees-of-freedom corrected partial $R^2$ (Geweke linear feedback \citep{geweke1982measurement}), which quantifies the linear directed predictability of one time series from another. As with TE, this per-pair $R^2$ is aggregated to a single dataset value by taking the median over all directed channel pairs, as reported in the main text.

\subsection{Transfer Entropy (TE)}

Now regarding the Transfer Entropy, we use the same joint differencing as Granger Causality, so the linear and non-linear axes see identical series. The main idea of TE is to measure the amount of directed information between two time series, capturing non-linear coupling that GC might miss. TE is computed based on KSG estimator \citep{kraskov2004estimating} quantifying the conditional mutual information between the present state of one time series and the past state of another, conditioned on the target's own most recent value. Conditioning on a single past value keeps the conditioning set low-dimensional, which the $k$-nearest-neighbor estimator requires to remain reliable. Computing the CMI on the ranks of the series (rather than their raw values) makes the estimate invariant to monotone transformations and robust to outliers, while the non-parametric estimator captures non-linear coupling of any form. Unlike GC, TE does not directly produce a p-value, so
we use a surrogate data approach to generate a null distribution of TE values. Following \citet{runge2018conditional}, we cast TE as the conditional independence test $X \perp Y \mid Z$ with source $X = x_{t-\ell}$, target $Y = y_t$, and conditioning set $Z = y_{t-1}$ (the target's own past). Rather than a full random shuffle, the surrogates are then drawn with a \emph{local-permutation} scheme: each source sample is permuted only among its $k_{\mathrm{perm}}$ nearest neighbors in the conditioning space $Z$, which breaks the $X \!\to\! Y$ information flow while preserving the marginal distributions and the dependence on $Z$. By comparing the observed TE value to this null distribution, we can assess the statistical significance of the detected information flow.

The test has a few essential parameters, which we set as follows:

\begin{itemize}
  \item \textbf{TE Max Lag ($\ell_{\max}^{\mathrm{TE}}$)}: We set the maximum lag $\ell_{\max}^{\mathrm{TE}}$ for TE computation to be equal to the smallest candidate lookback window of the dataset (the order $p$ of GC). This is done to make sure that every coupling identified by TE is relevant to the forecasting task and not just a coupling in a lag that the model is not able to use.
  \item \textbf{Number of Surrogates}: We generate 100 surrogate time series for each pair of time series to create a robust null distribution. This number is taken directly from the literature \citep{runge2018conditional} and is a good trade-off between computational cost and statistical power (as a higher number of surrogates does not yield higher stability in results anymore).
  \item \textbf{$k_{\mathrm{CMI}}$}: The number of nearest neighbors used in the KSG estimator of the CMI. This value also followed the original paper's rule of thumb of picking the $k$ value based on the number of usable samples $n$ in the dataset, $k_{\mathrm{CMI}}=\min(0.2\,n, 200)$. Scaling $k$ with $n$ lowers the estimator's variance on long series, and the cap bounds the computational cost.
  \item \textbf{$k_{\mathrm{perm}}$}: The number of nearest neighbors (in the conditioning space $Z$) used by the local-permutation scheme for the surrogate null distribution. This is set universally for all datasets as 5, which was shown to work well for a variety of the datasets in the original paper \citep{runge2018conditional}.
  \item \textbf{Lag Sweep \& Peak Selection ($\ell^\star$)}: For each ordered pair we evaluate the effective TE at every candidate lag $\ell \in \{1,\dots,\ell_{\max}^{\mathrm{TE}}\}$. The reported value is the \emph{effective} (debiased) transfer entropy $\mathrm{TE}^{\mathrm{eff}}=\max\!\bigl(0,\,\widehat{I}(y_t;x_{t-\ell}\mid y_{t-1})-\overline{I}_{\text{null}}\bigr)$, which subtracts the surrogate-null mean $\overline{I}_{\text{null}}$ from the raw KSG estimate (the same debiasing applied to MI). We test each swept lag against its surrogate null and apply a Benjamini--Hochberg false-discovery-rate correction across the swept lags \citep{benjamini1995controlling}. This controls the expected proportion of lags falsely flagged as significant. The reported lag $\ell^\star$ is the FDR-significant lag with the largest effective TE, and the effective TE is set to $0$ if no lag survives the correction. This is the sweep procedure referenced from the main text (Section~\ref{sec:te}).
\end{itemize}

Finally, based on the above preprocessing and parameter settings, we compute the TE values for all pairs of channels in each dataset and aggregate them into the median TE $R^2$ used in the main text (results in Appendix~\ref{sec:stat-profiling-more-results}).

\subsection{Mutual Information (MI)}

Now regarding the Mutual Information, unlike GC and TE it does not condition on the target's own past. At lag $0$, it is a symmetric measure of the overall coupling shared between a pair of channels. At positive lags, it is directional, from source to target, as in GC and TE. It also differs in preprocessing: MI is estimated on the \emph{raw} series (rank-transformed for scale invariance, \emph{without} the seasonal and regular differencing applied for GC and TE), so that shared trend and seasonality are retained as legitimate common-driver information rather than removed. The estimator itself is the same KSG estimator used for TE \citep{kraskov2004estimating}, and, as with TE, every MI estimate is debiased against a surrogate null (see the \textbf{Number of Surrogates} parameter below). We mainly use MI to estimate whether cross-channel coupling is lagged or contemporaneous. We set its parameters as follows:

\begin{itemize}
  \item \textbf{$k_{\mathrm{MI}}$}: The number of nearest neighbors used in the KSG estimator for MI computation. We set this value to 5, a common choice for the KSG estimator \citep{kraskov2004estimating}.
  \item \textbf{Number of Surrogates}: Similar to TE, we generate 100 surrogate time series for each pair of channels to create a null distribution for the MI values, chosen at this value for consistency with TE. The surrogates are drawn by randomly permuting the source series, which destroys any coupling with the target while preserving its marginal distribution. Every reported MI value --- both contemporaneous and lagged --- is \emph{effective}, i.e.\ debiased by subtracting the mean of this surrogate null from the raw KSG estimate. The peak-lag and lag-$0$ quantities that enter $\phi_{\mathrm{lag}}$ are thus compared on the same debiased scale.
  \item \textbf{Lag Grid}: We evaluate MI in two setups, contemporaneous and lagged. The contemporaneous MI is computed between the present states of the two channels ($\ell = 0$), while the lagged MI is evaluated over the geometrically spaced candidate lags in $\Lambda$ (Eq.~\ref{eq:mi-lags}) and retains the positive lag $\ell^\star \in \Lambda$ that maximizes the effective MI. This is the lag reported in the main text. The selection is thus driven by the MI values on this fixed lag grid, not by any cross-correlation criterion. Here is how the lag grid is defined for each dataset:
  \begin{equation}
  \Lambda \;=\; \{0\}\;\cup\;
  \Bigl\{\, \bigl\lfloor \ell_{\mathrm{top}}^{\,i/(m-1)} \bigr\rceil \;:\; i = 0,1,\dots,m-1 \Bigr\},
  \qquad
  \ell_{\mathrm{top}} \;=\; \min\!\bigl(\ell_{\max},\,\ell_{\mathrm{cap}}\bigr),
  \label{eq:mi-lags}
\end{equation}
where $m = 6$ is the number of positive-lag grid points, $\lfloor\cdot\rceil$
denotes rounding to the nearest integer, $\ell_{\mathrm{cap}} = 300$ caps the
largest lag, and $\ell_{\max}$ is the dataset-specific lag window. The resulting grid of each dataset is listed in Table~\ref{tab:mi_lag_grid}. Duplicate integers introduced by rounding are collapsed, so $\Lambda$ is a
set. The geometric spacing concentrates grid points at short lags---where
cross-channel coupling typically resides---while still reaching $\ell_{\mathrm{top}}$. For example,
$\ell_{\mathrm{top}} = 300$ yields $\Lambda = \{0,1,3,10,31,96,300\}$ and $\ell_{\mathrm{top}} = 48$ yields
$\Lambda = \{0,1,2,5,10,22,48\}$.
\end{itemize}

\begin{table}[h]
\centering\small
\caption{Lag grid of the lagged-MI measure. For each dataset, MI is evaluated at lag 0 and at six log-spaced lags in $[1, \ell_{\mathrm{top}}]$, where $\ell_{\mathrm{top}}=\min(\ell_{\max},\ell_{\mathrm{cap}})$ is the dataset's lag window capped at $\ell_{\mathrm{cap}}=300$ steps ($^\dagger$ = capped). The synthetic grid is used for profiling only, while the retrieval in Section~\ref{sec:statistical-methods-validation} uses the dense grid $\Lambda_{\mathrm{syn}}$.}
\label{tab:mi_lag_grid}
\setlength{\tabcolsep}{4pt}
\begin{tabular}{llrll}
\toprule
Dataset(s) & Sampling & $\ell_{\mathrm{top}}$ & Lag grid (steps) & Span of $\ell_{\mathrm{top}}$ \\
\midrule
\multicolumn{5}{l}{\textit{Standard real-world}} \\
ETTh1, ETTh2, Electricity, Traffic & 1\,h     & 168 & 1, 3, 8, 22, 60, 168 & 1 week \\
ETTm1, ETTm2                       & 15\,min   & 300$^\dagger$ & 1, 3, 10, 31, 96, 300 & 3.1 days \\
Weather, SolarAL                   & 10\,min   & 168 & 1, 3, 8, 22, 60, 168 & 28\,h \\
ExchangeRate                       & 1 day     & 60  & 1, 2, 5, 12, 26, 60 & 2 months \\
ILI                                & 1 week    & 52  & 1, 2, 5, 11, 24, 52 & 1 year \\
\midrule
\multicolumn{5}{l}{\textit{Spatio-temporal (traffic)}} \\
METR-LA, PEMS-BAY                  & 5\,min    & 288 & 1, 3, 10, 30, 93, 288 & 1 day \\
\midrule
\multicolumn{5}{l}{\textit{Coupled real-world}} \\
AirQuality, AQShunyi               & 1\,h      & 168 & 1, 3, 8, 22, 60, 168 & 1 week \\
CzeLan, ZafNoo                     & 30\,min   & 300$^\dagger$ & 1, 3, 10, 31, 96, 300 & 6.25 days \\
Bavaria-Iller, Bavaria-Donau,      & 15\,min   & 300$^\dagger$ & 1, 3, 10, 31, 96, 300 & 3.1 days \\
\quad HouseholdPower               &           &     &                      &          \\
WaterQuality-Darwin                & 15\,min   & 192 & 1, 3, 8, 23, 67, 192 & 2 days \\
Current-Velocity                   & 20\,min   & 144 & 1, 3, 7, 20, 53, 144 & 2 days \\
Wearable                           & 5\,min    & 288 & 1, 3, 10, 30, 93, 288 & 1 day \\
\midrule
Chaotic ODE systems                & $\Delta t=0.05$ & 50  & 1, 2, 5, 10, 23, 50 & 2.5 time units \\
Synthetic datasets                 & 1 step    & 168 & 1, 3, 8, 22, 60, 168 & 168 steps \\
\bottomrule
\end{tabular}
\end{table}

Finally, based on the above preprocessing and parameter settings, we compute the contemporaneous and lagged MI for all pairs of channels in each dataset. These are aggregated into the \emph{lagged-coupling fraction} $\phi_{\mathrm{lag}}$ defined in the main text, the fraction of pairs whose coupling is stronger at the peak lag $\ell^\star$ than at lag $0$, since it directly separates lagged transfer from contemporaneous (potentially redundant) coupling.

\section{Hyperparameter Ranges}
\label{sec:hyperparameter-tuning-setup}

Every (model, dataset, horizon) combination is tuned separately with Optuna's TPE
sampler (20 trials), selecting the trial with the lowest validation MSE. All
trials share the same training protocol: Adam, at most 30 epochs with early
stopping on the validation MSE (patience 5), and a learning rate halved after
every epoch. Table~\ref{tab:hp_ranges} lists the tuned hyperparameters of each
model. Ranges that depend on the number of channels $C$ of a dataset, which bound
the memory footprint on wide datasets, are given in
Table~\ref{tab:hp_channel_ranges}. The lookback window is tuned jointly with these
hyperparameters, and its candidates are listed per dataset in
Appendix~\ref{sec:datasets-details}.

\begin{table}[h]
\centering\small
\caption{Tuned hyperparameters and their search ranges. $\mathcal{B}$, $\mathcal{D}$,
$\mathcal{H}$ and $N_{\max}$ are the channel-dependent ranges of
Table~\ref{tab:hp_channel_ranges}. $\log\mathcal{U}$ denotes log-uniform and
$\mathcal{U}$ uniform sampling.}
\label{tab:hp_ranges}
\begin{tabular}{lll}
\toprule
Model & Hyperparameter & Range \\
\midrule
All except CycleNet & learning rate & $\log\mathcal{U}[10^{-4}, 10^{-2}]$ \\
All models          & batch size & $\mathcal{B}$ \\
\midrule
TSMixer, TSMixer\textsubscript{CI}
  & number of mixer blocks & $\{1,\dots,N_{\max}\}$ \\
  & hidden size of the mixing MLPs & $\mathcal{H}$ \\
  & dropout & $\mathcal{U}[0.01, 0.3]$ \\
  & per-channel output head & $\{\text{yes}, \text{no}\}$ \\
\midrule
DLinear & per-channel linear layers & $\{\text{yes}, \text{no}\}$ \\
\midrule
CycleNet & learning rate & $\{2\cdot10^{-3}, 5\cdot10^{-3}, 10^{-2}\}$ \\
         & backbone & $\{\text{linear}, \text{MLP}\}$ \\
\midrule
iTransformer & model dimension $d_{\text{model}}$ & $\mathcal{D}$ \\
             & attention heads & $\{4, 8, 16\}$ \\
             & encoder layers & $\{1, 2, 3\}$ \\
             & dropout & $\mathcal{U}[0.01, 0.3]$ \\
\midrule
SimpleTM & model dimension $d_{\text{model}}$ & $\mathcal{D}$ \\
         & encoder layers & $\{1, 2, 3\}$ \\
         & dropout & $\mathcal{U}[0.01, 0.3]$ \\
         & wavelet & $\{\texttt{db1}, \texttt{bior3.1}, \texttt{db4}\}$ \\
         & wavelet decomposition levels $m$ & $\{1, 2, 3\}$ \\
         & geometric-attention weight $\alpha$ & $\mathcal{U}[0, 1]$ \\
         & geometric-attention dropout & $\mathcal{U}[0.1, 0.5]$ \\
         & L1 attention regularization weight & $\{0, 5\cdot10^{-5}, 5\cdot10^{-4}, 5\cdot10^{-3}\}$ \\
\bottomrule
\end{tabular}
\end{table}

\begin{table}[h]
\centering\small
\caption{Channel-dependent search ranges ($C$ = number of channels).}
\label{tab:hp_channel_ranges}
\setlength{\tabcolsep}{4pt}
\begin{tabular}{lcccc}
\toprule
 & $C \le 50$ & $50 < C \le 200$ & $200 < C \le 500$ & $C > 500$ \\
\midrule
batch size $\mathcal{B}$              & $\{16,32,64,128\}$ & $\{16,32,64\}$ & $\{8,16,32\}$ & $\{4,8,16\}$ \\
$d_{\text{model}}$ $\mathcal{D}$       & $\{64,128,256,512\}$ & $\{64,128,256\}$ & $\{64,128,256\}$ & $\{64,128\}$ \\
TSMixer hidden size $\mathcal{H}$     & $\{32,64,128,256,512\}$ & $\{32,64,128,256\}$ & $\{32,64,128\}$ & $\{32,64,128\}$ \\
TSMixer blocks $N_{\max}$             & 6 & 4 & 3 & 3 \\
\bottomrule
\end{tabular}
\end{table}

Some settings are fixed rather than tuned. The feed-forward width of the
transformers is $4\,d_{\text{model}}$, and the DLinear moving-average kernel is 25. The
CycleNet MLP backbone has hidden size 512 and uses RevIN without affine parameters,
and its cycle length is set per dataset to the dominant seasonal period.
TSMixer\textsubscript{CI} shares the search space of TSMixer, with the
channel-mixing MLP disabled.

\section{Statistical Profiling - More Results}
\label{sec:stat-profiling-more-results}

Table~\ref{tab:gc_te_results} reports GC and TE, the two statistical measures that do not retrieve all planted couplings in Section~\ref{sec:measurement-evaluation}, next to \milag{} and \cdg{} for all datasets. At the suite level, both measures are higher on the coupled datasets than on the standard datasets. The median $\widetilde{R^2_{\mathrm{GC}}}$ is $0.015$ vs.\ $0.005$ (about $3\times$), and the median $\widetilde{R^2_{\mathrm{TE}}}$ is $0.019$ vs.\ $0.009$ (about $2.1\times$). TE exceeds the standard median on 9 of the 10 coupled datasets, and GC on 8 of 10.

\begin{figure}[h]
\centering
\begin{subfigure}[t]{0.49\linewidth}\centering
  \includegraphics[width=\linewidth]{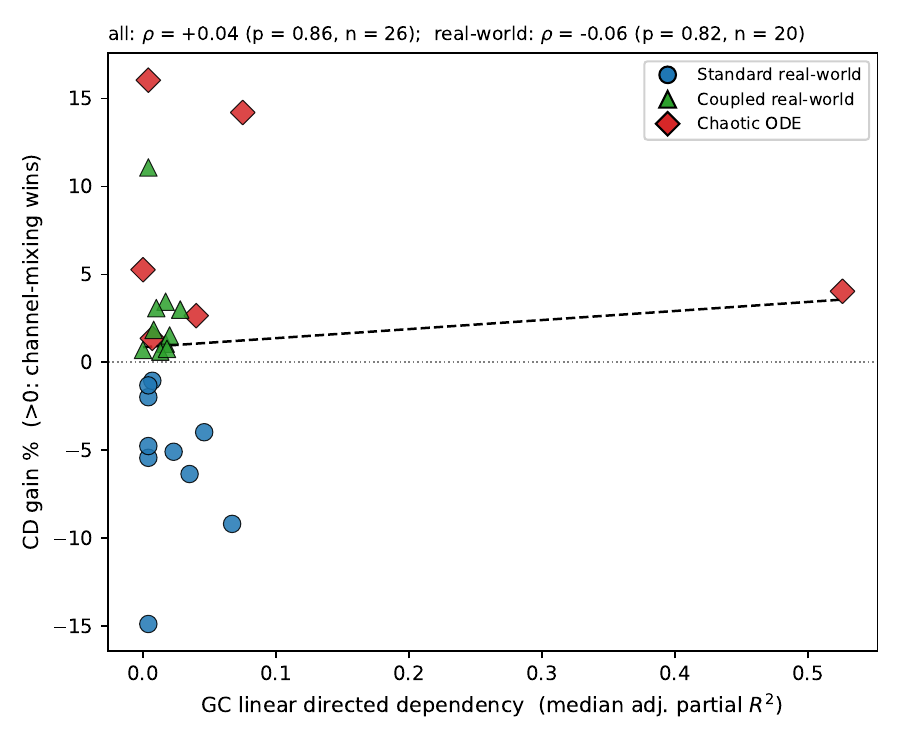}
  \caption{GC}
\end{subfigure}\hfill
\begin{subfigure}[t]{0.49\linewidth}\centering
  \includegraphics[width=\linewidth]{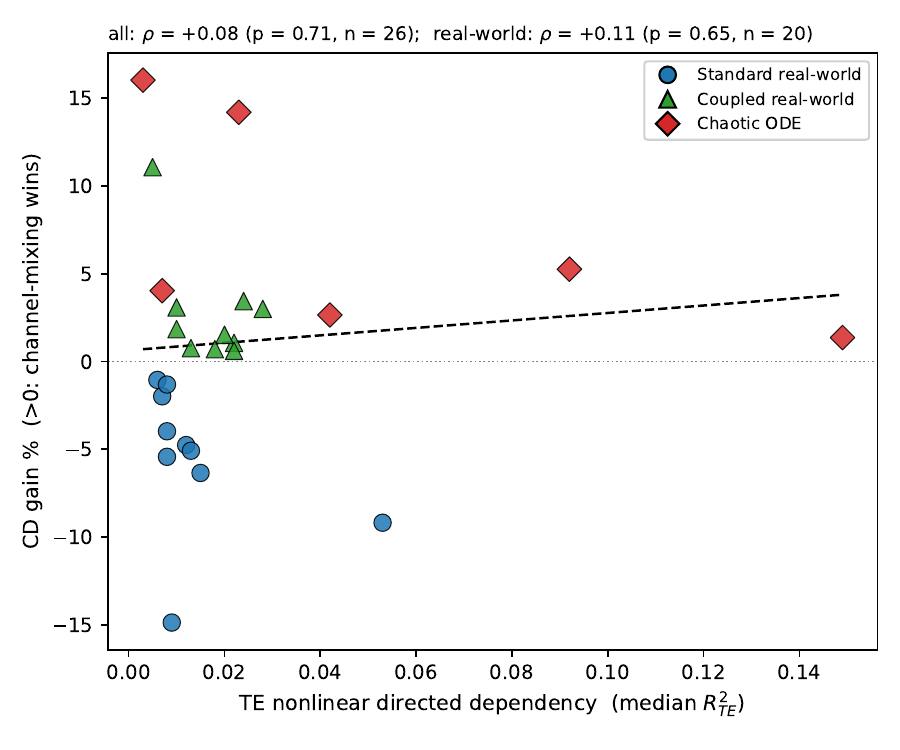}
  \caption{TE}
\end{subfigure}
\caption{Median GC (a) and TE (b) vs.\ \cdg{} (\%) of TSMixer over TSMixer\textsubscript{CI} on 26 datasets (10 standard, 10 coupled, 6 ODE). The dashed line is a linear fit, and Spearman correlations are given above each plot. Compare with \milag{} in Figure~\ref{fig:motivation}.}
\label{fig:gc_te_cdg}
\end{figure}

At the dataset level, however, neither measure reliably separates the two suites or tracks the channel-mixing payoff. Traffic has the highest GC ($0.067$) and TE ($0.053$) of all real-world datasets, yet one of the most negative \cdg{} values ($-9.2\%$). Conversely, Bavaria-Donau, the coupled dataset with the largest \cdg{} ($+11.1\%$), has among the lowest GC and TE values ($0.004$ and $0.005$). The same holds for the ODE suite, where Blinking-Rotlet combines the largest \cdg{} ($+16.0\%$) with GC and TE values close to zero. Across all 26 datasets, the Spearman correlation with \cdg{} is only $0.04$ for GC and $0.08$ for TE (Figure~\ref{fig:gc_te_cdg}), compared to $0.78$ for \milag{} (Figure~\ref{fig:motivation}). This matches the synthetic results, where GC misses the non-linear and joint couplings and TE mis-locates the contemporaneous coupling (Table~\ref{tab:synth_retrieval_detailed}). The magnitude of GC and TE therefore reflects the strength of a specific functional form rather than whether channel mixing pays off, which is why we use \milag{} and \cdg{} in the main analysis.

\begin{table}[h]
\centering
\small
\caption{Median GC and TE over all ordered channel pairs ($\widetilde{R^2_{\mathrm{GC}}}$ and $\widetilde{R^2_{\mathrm{TE}}}$, Section~\ref{sec:statistical-profiling}), next to \milag{} and \cdg{} as in Tables~\ref{tab:analysis_standard} and~\ref{tab:occasional_results}.}
\label{tab:gc_te_results}
\begin{tabular}{l cc cc}
\toprule
Dataset & $\widetilde{R^2_{\mathrm{GC}}}$ & $\widetilde{R^2_{\mathrm{TE}}}$ & \milag{} (\%) & \cdg{} (\%) \\
\midrule
\multicolumn{5}{l}{\textit{Standard MTSF}} \\
\quad ETTh1 & 0.004 & 0.009 & 19 & $-14.9$ \\
\quad ETTh2 & 0.004 & 0.008 & 19 & $-5.4$ \\
\quad ETTm1 & 0.004 & 0.012 & 38 & $-4.8$ \\
\quad ETTm2 & 0.004 & 0.007 & 38 & $-2.0$ \\
\quad Weather & 0.007 & 0.006 & 52 & $-1.0$ \\
\quad Electricity & 0.023 & 0.013 & 61 & $-5.1$ \\
\quad Traffic & 0.067 & 0.053 & 27 & $-9.2$ \\
\quad ExchangeRate & 0.004 & 0.008 & 7 & $-1.3$ \\
\quad ILI & 0.035 & 0.015 & 17 & $-6.3$ \\
\quad SolarAL & 0.046 & 0.008 & 14 & $-4.0$ \\
\cmidrule(l){2-5}
\quad \textit{category median} & \textit{0.005} & \textit{0.009} & \textit{23} & $\mathit{-4.9}$ \\
\midrule
\multicolumn{5}{l}{\textit{Coupled}} \\
\quad Bavaria-Donau & 0.004 & 0.005 & 78 & $+11.1$ \\
\quad Bavaria-Iller & 0.017 & 0.024 & 67 & $+3.4$ \\
\quad Current-Velocity & 0.010 & 0.010 & 83 & $+3.1$ \\
\quad CzeLan & 0.028 & 0.028 & 82 & $+3.0$ \\
\quad ZafNoo & 0.008 & 0.010 & 57 & $+1.8$ \\
\quad AirQuality & 0.020 & 0.020 & 54 & $+1.5$ \\
\quad AQShunyi & 0.017 & 0.022 & 53 & $+1.1$ \\
\quad WaterQuality-Darwin & 0.018 & 0.013 & 40 & $+0.8$ \\
\quad Wearable & 0.000 & 0.018 & 17 & $+0.7$ \\
\quad HouseholdPower & 0.013 & 0.022 & 17 & $+0.6$ \\
\cmidrule(l){2-5}
\quad \textit{category median} & \textit{0.015} & \textit{0.019} & \textit{55.5} & $\mathit{+1.7}$ \\
\midrule
\multicolumn{5}{l}{\textit{Spatio-temporal (traffic)}} \\
\quad METR-LA & 0.011 & 0.025 & 38 & $-1.5$ \\
\quad PEMS-BAY & 0.021 & 0.015 & 69 & $+6.4$ \\
\cmidrule(l){2-5}
\quad \textit{category median} & \textit{0.016} & \textit{0.020} & \textit{53.5} & $\mathit{+2.4}$ \\
\midrule
\multicolumn{5}{l}{\textit{Chaotic ODE (known-correlated reference)}} \\
\quad Lorenz & 0.007 & 0.149 & 83 & $+1.4$ \\
\quad Lorenz-Coupled & 0.000 & 0.092 & 63 & $+5.3$ \\
\quad Blinking-Rotlet & 0.004 & 0.003 & 83 & $+16.0$ \\
\quad CellCycle & 0.040 & 0.042 & 67 & $+2.7$ \\
\quad Double-Pendulum & 0.526 & 0.007 & 92 & $+4.0$ \\
\quad Hopfield & 0.075 & 0.023 & 73 & $+14.2$ \\
\cmidrule(l){2-5}
\quad \textit{category median} & \textit{0.024} & \textit{0.033} & \textit{78} & $\mathit{+4.6}$ \\
\bottomrule
\end{tabular}
\end{table}

\section{Parameter Sensitivity}
\label{sec:parameter-sensitivity}

Pairwise profiling scales quadratically with the number of channels, so datasets
with more than 40 channels are profiled on a random subset of $|\mathcal{C}|=40$ channels,
drawn uniformly without replacement with a fixed seed (2021). All $|\mathcal{P}|=|\mathcal{C}|(|\mathcal{C}|-1)$ ordered
pairs of the subset are analyzed. We test whether this choice affects the
dataset-level dependency measures on the two widest datasets with a completed
sweep, Electricity (321 channels) and PEMS-BAY (325 channels). Following a
one-factor-at-a-time design, we vary the subset size
$|\mathcal{C}|\in\{20,40,80\}$ at the default seed, and the sampling seed over
$\{2021,7,13\}$ at the default size $|\mathcal{C}|=40$ (for PEMS-BAY also seed 7 at
$|\mathcal{C}|=20$). All other settings (estimators, lag grids and deseasonalization) are those of the main analysis.
Table~\ref{tab:channel_sampling} reports the median degrees-of-freedom corrected partial $R^2$ of GC,
$\widetilde{R^2_{\mathrm{GC}}}$, the median TE $R^2$, $\widetilde{R^2_{\mathrm{TE}}}$, and the lagged-coupling fraction $\phi_{\mathrm{lag}}$ of \milag{}, all computed over the analyzed pairs (Section~\ref{sec:statistical-profiling}).

\begin{table}[h]
\centering\small\setlength{\tabcolsep}{3.5pt}
\caption{Channel-subsampling sensitivity of $\widetilde{R^2_{\mathrm{GC}}}$ (GC), $\widetilde{R^2_{\mathrm{TE}}}$ (TE), and \milag{} over all ordered pairs of the subset, for subset size $|\mathcal{C}|$ and sampling seed $s$. $\Delta$: deviation from the default setting ($|\mathcal{C}|=40$, $s=2021$, marked $^\ast$), relative for GC and TE and in percentage points (pp) for \milag{}.}
\label{tab:channel_sampling}
\begin{tabular}{llrrrrrrr}
\toprule
Dataset & Setting & Pairs & GC & $\Delta$GC & TE & $\Delta$TE & \milag{} (\%) & $\Delta$\milag{} (pp) \\
\midrule
Electricity & $|\mathcal{C}|=20$, $s=2021$ &   380 & 0.0215 & $-4.8\%$ & 0.0129 & $-3.7\%$ & 66.6 & $+5.2$ \\
            & $|\mathcal{C}|=40$, $s=2021^\ast$ & 1560 & 0.0225 & --- & 0.0134 & --- & 61.4 & --- \\
            & $|\mathcal{C}|=80$, $s=2021$ &  6320 & 0.0210 & $-6.7\%$ & 0.0137 & $+2.2\%$ & 62.9 & $+1.5$ \\
            & $|\mathcal{C}|=40$, $s=7$    &  1560 & 0.0247 & $+9.5\%$ & 0.0147 & $+10.1\%$ & 67.2 & $+5.8$ \\
            & $|\mathcal{C}|=40$, $s=13$   &  1560 & 0.0204 & $-9.7\%$ & 0.0125 & $-6.7\%$ & 68.4 & $+7.0$ \\
\midrule
PEMS-BAY    & $|\mathcal{C}|=20$, $s=2021$ &   380 & 0.0248 & $+20.1\%$ & 0.0152 & $+1.3\%$ & 64.2 & $-4.5$ \\
            & $|\mathcal{C}|=20$, $s=7$    &   380 & 0.0170 & $-17.3\%$ & 0.0124 & $-17.0\%$ & 74.2 & $+5.5$ \\
            & $|\mathcal{C}|=40$, $s=2021^\ast$ & 1560 & 0.0206 & --- & 0.0150 & --- & 68.7 & --- \\
            & $|\mathcal{C}|=80$, $s=2021$ &  6320 & 0.0188 & $-8.8\%$ & 0.0143 & $-4.7\%$ & 67.5 & $-1.2$ \\
            & $|\mathcal{C}|=40$, $s=7$    &  1560 & 0.0193 & $-6.2\%$ & 0.0142 & $-5.3\%$ & 67.2 & $-1.5$ \\
            & $|\mathcal{C}|=40$, $s=13$   &  1560 & 0.0151 & $-26.9\%$ & 0.0142 & $-5.3\%$ & 69.7 & $+1.0$ \\
\bottomrule
\end{tabular}
\end{table}

\paragraph{Number of sampled channels.}
Doubling the subset from $|\mathcal{C}|=40$ to $|\mathcal{C}|=80$ (four times as many pairs) changes the
medians by at most $9\%$ (GC $-6.7\%$ and $-8.8\%$, TE $+2.2\%$ and $-4.7\%$ on
Electricity and PEMS-BAY). Halving it to $|\mathcal{C}|=20$ is less stable: with only 380
pairs, the PEMS-BAY GC median deviates by $+20\%$ at seed 2021 and by $-17\%$ at
seed 7. \milag{} is similarly stable. It moves by $+1.5$ and $-1.2$ pp at $|\mathcal{C}|=80$ and by at most $5.5$ pp at $|\mathcal{C}|=20$. A cap of $|\mathcal{C}|=40$ is therefore large enough for the medians to settle,
while keeping the pairwise TE tests tractable.

\paragraph{Sampling seed.}
At $|\mathcal{C}|=40$, the choice of subset moves the Electricity medians by up to
$\pm10\%$ (coefficient of variation across seeds: $9.6\%$ for GC, $8.4\%$ for TE).
On PEMS-BAY, TE is stable (CV $3.2\%$), whereas GC varies more (CV $15.8\%$, with seed
13 giving $-27\%$), reflecting the heterogeneity of its sensor network. \milag{} shifts by up to $+7.0$ pp on Electricity and by at most $1.5$ pp on PEMS-BAY.

\paragraph{Effect on the cross-dataset comparison.}
These deviations are small relative to the differences between datasets, whose
median GC effect sizes span more than an order of magnitude (from about $0.004$ on
the ETT datasets to $0.067$ on Traffic) and whose \milag{} values range from $7\%$ (ExchangeRate) to $83\%$ (Current-Velocity). In particular, Electricity keeps the same
rank among the profiled real-world datasets on GC and TE under all five
settings. On \milag{}, both datasets stay above $50\%$ lagged-coupled pairs under all settings ($61$--$68\%$ on Electricity and $64$--$74\%$ on PEMS-BAY), and Electricity remains the standard dataset with the highest \milag{}. The subsampling therefore does not change the dataset ordering on which
the coupling analysis relies.

\section{Seed-Averaged Results and Standard Deviations}
\label{sec:random-seed-full-results}

\begin{table*}[t]
\centering\small
\caption{Seed-averaged test MSE, i.e., the mean over seeds 2021/2025/2026 of the horizon-averaged MSE. Best per row in \textbf{bold} and second \underline{underlined}, with ties all bold and Win listing the strategies. Models grouped by channel strategy and shaded: \colorbox{ciband}{\catCI} channel-independent (TSM\textsubscript{CI}=TSMixer\textsubscript{CI}, DLin.=DLinear, CycNet=CycleNet), \colorbox{cdband}{\catCD} channel-dependent (TSM=TSMixer, SimTM=SimpleTM, iTr.=iTransformer).}
\label{tab:seed_mse_mean}
\begin{tabular}{l >{\columncolor{ciband}}c >{\columncolor{ciband}}c >{\columncolor{ciband}}c >{\columncolor{cdband}}c >{\columncolor{cdband}}c >{\columncolor{cdband}}c c}
\toprule
  & \multicolumn{3}{c}{\cellcolor{ciband}\textbf{CI}} & \multicolumn{3}{c}{\cellcolor{cdband}\textbf{CD}} &   \\
\cmidrule(lr){2-4}\cmidrule(lr){5-7}
Dataset & TSM\textsubscript{CI} & DLin. & CycNet & TSM & SimTM & iTr. & Win \\
\midrule
\multicolumn{8}{l}{\textit{Standard MTSF}} \\
ETTh1 & 0.433 & 0.456 & \textbf{0.413} & 0.498 & \underline{0.427} & 0.451 & \catCI \\
ETTh2 & 0.390 & 0.441 & \textbf{0.370} & 0.411 & \underline{0.379} & 0.382 & \catCI \\
ETTm1 & \textbf{0.352} & 0.359 & \underline{0.354} & 0.369 & 0.357 & 0.367 & \catCI \\
ETTm2 & 0.266 & \underline{0.263} & \textbf{0.249} & 0.271 & 0.265 & 0.273 & \catCI \\
Weather & \underline{0.232} & \textbf{0.222} & \underline{0.232} & 0.235 & 0.235 & 0.238 & \catCI \\
Electricity & \textbf{0.158} & 0.161 & \underline{0.159} & 0.166 & 0.162 & 0.162 & \catCI \\
Traffic & \textbf{0.391} & 0.410 & 0.408 & 0.427 & \underline{0.399} & 0.418 & \catCI \\
ExchangeRate & \underline{0.152} & \textbf{0.128} & 0.173 & 0.154 & 0.182 & 0.165 & \catCI \\
ILI & \textbf{1.873} & 2.388 & 1.952 & 1.992 & \underline{1.900} & 2.382 & \catCI \\
SolarAL & \textbf{0.193} & 0.224 & 0.202 & 0.200 & 0.200 & \underline{0.198} & \catCI \\
\midrule
\multicolumn{8}{l}{\textit{Spatio-temporal (traffic)}} \\
METR-LA & \underline{0.916} & \textbf{0.898} & 0.919 & 0.930 & 0.951 & 1.020 & \catCI \\
PEMS-BAY & 0.501 & 0.613 & 0.528 & 0.469 & \textbf{0.423} & \underline{0.446} & \catCD \\
\midrule
\multicolumn{8}{l}{\textit{Coupled}} \\
AirQuality & 0.750 & \textbf{0.738} & 0.744 & \underline{0.739} & 0.752 & 0.748 & \catCI \\
AQShunyi & 0.684 & \underline{0.672} & 0.682 & 0.677 & \textbf{0.664} & 0.697 & \catCD \\
CzeLan & 0.212 & 0.218 & 0.216 & \textbf{0.205} & \underline{0.211} & 0.225 & \catCD \\
ZafNoo & 0.523 & \textbf{0.486} & 0.528 & \underline{0.514} & 0.529 & 0.539 & \catCI \\
Bavaria-Iller & 0.185 & \underline{0.176} & 0.186 & 0.179 & \textbf{0.170} & 0.186 & \catCD \\
Bavaria-Donau & 0.097 & \underline{0.095} & 0.098 & \textbf{0.086} & \textbf{0.086} & 0.100 & \catCD \\
WaterQuality-Darwin & \underline{0.270} & 0.281 & 0.271 & \textbf{0.268} & 0.275 & 0.280 & \catCD \\
Current-Velocity & 0.263 & 0.264 & 0.270 & \underline{0.255} & \textbf{0.254} & 0.265 & \catCD \\
Wearable & 0.671 & \underline{0.668} & 0.781 & \textbf{0.666} & 0.753 & 0.755 & \catCD \\
HouseholdPower & 0.652 & \textbf{0.644} & \underline{0.646} & 0.648 & 0.648 & \underline{0.646} & \catCI \\
\bottomrule
\end{tabular}
\end{table*}

\begin{table*}[t]
\centering\small
\caption{Seed-averaged test MAE, i.e., the mean over seeds 2021/2025/2026 of the horizon-averaged MAE. Best per row in \textbf{bold} and second \underline{underlined}, with ties all bold and Win listing the strategies. Models grouped by channel strategy and shaded: \colorbox{ciband}{\catCI} channel-independent (TSM\textsubscript{CI}=TSMixer\textsubscript{CI}, DLin.=DLinear, CycNet=CycleNet), \colorbox{cdband}{\catCD} channel-dependent (TSM=TSMixer, SimTM=SimpleTM, iTr.=iTransformer).}
\label{tab:seed_mae_mean}
\begin{tabular}{l >{\columncolor{ciband}}c >{\columncolor{ciband}}c >{\columncolor{ciband}}c >{\columncolor{cdband}}c >{\columncolor{cdband}}c >{\columncolor{cdband}}c c}
\toprule
  & \multicolumn{3}{c}{\cellcolor{ciband}\textbf{CI}} & \multicolumn{3}{c}{\cellcolor{cdband}\textbf{CD}} &   \\
\cmidrule(lr){2-4}\cmidrule(lr){5-7}
Dataset & TSM\textsubscript{CI} & DLin. & CycNet & TSM & SimTM & iTr. & Win \\
\midrule
\multicolumn{8}{l}{\textit{Standard MTSF}} \\
ETTh1 & \underline{0.437} & 0.457 & \textbf{0.425} & 0.473 & 0.440 & 0.459 & \catCI \\
ETTh2 & 0.408 & 0.449 & \textbf{0.401} & 0.422 & \underline{0.406} & \underline{0.406} & \catCI \\
ETTm1 & \textbf{0.380} & 0.382 & \underline{0.381} & 0.392 & 0.383 & 0.394 & \catCI \\
ETTm2 & \underline{0.320} & 0.330 & \textbf{0.313} & 0.325 & 0.321 & 0.330 & \catCI \\
Weather & \underline{0.271} & 0.279 & \underline{0.271} & 0.272 & \textbf{0.270} & 0.275 & \catCD \\
Electricity & \textbf{0.251} & 0.258 & \underline{0.252} & 0.266 & 0.256 & 0.258 & \catCI \\
Traffic & \textbf{0.268} & 0.282 & \underline{0.278} & 0.306 & 0.282 & 0.298 & \catCI \\
ExchangeRate & \underline{0.253} & \textbf{0.246} & 0.267 & 0.255 & 0.273 & 0.266 & \catCI \\
ILI & \textbf{0.902} & 1.095 & 0.932 & \underline{0.909} & \textbf{0.902} & 1.027 & \catCD/\catCI \\
SolarAL & \textbf{0.240} & 0.287 & \underline{0.253} & 0.261 & 0.257 & 0.261 & \catCI \\
\midrule
\multicolumn{8}{l}{\textit{Spatio-temporal (traffic)}} \\
METR-LA & 0.555 & 0.587 & 0.551 & \underline{0.550} & \textbf{0.546} & 0.580 & \catCD \\
PEMS-BAY & 0.338 & 0.401 & 0.352 & 0.323 & \textbf{0.298} & \underline{0.300} & \catCD \\\midrule
\multicolumn{8}{l}{\textit{Coupled}} \\
AirQuality & 0.502 & 0.504 & \underline{0.495} & \textbf{0.494} & 0.496 & 0.499 & \catCD \\
AQShunyi & 0.507 & 0.508 & 0.505 & \underline{0.504} & \textbf{0.497} & 0.510 & \catCD \\
CzeLan & \textbf{0.256} & 0.277 & 0.264 & \underline{0.259} & 0.262 & 0.279 & \catCI \\
ZafNoo & 0.462 & \textbf{0.440} & 0.461 & \underline{0.459} & 0.470 & 0.472 & \catCI \\
Bavaria-Iller & 0.187 & 0.207 & 0.188 & \underline{0.184} & \textbf{0.182} & 0.187 & \catCD \\
Bavaria-Donau & 0.169 & 0.168 & 0.173 & \textbf{0.155} & \underline{0.156} & 0.172 & \catCD \\
WaterQuality-Darwin & 0.188 & 0.190 & 0.190 & \textbf{0.185} & \underline{0.187} & 0.190 & \catCD \\
Current-Velocity & 0.269 & 0.271 & 0.274 & \underline{0.263} & \textbf{0.262} & 0.271 & \catCD \\
Wearable & 0.592 & \textbf{0.590} & 0.599 & \textbf{0.590} & \underline{0.591} & 0.594 & \catCD/\catCI \\
HouseholdPower & 0.509 & 0.514 & \underline{0.505} & 0.508 & \textbf{0.504} & \textbf{0.504} & \catCD \\
\bottomrule
\end{tabular}
\end{table*}

\begin{table*}[t]
\centering\small
\caption{Across-seed std of the horizon-averaged test MSE (seeds 2021/2025/2026), reported to gauge significance. Columns: TSM\textsubscript{CI}=TSMixer\textsubscript{CI}, DLin.=DLinear, CycNet=CycleNet (CI), and TSM=TSMixer, SimTM=SimpleTM, iTr.=iTransformer (CD).}
\label{tab:seed_mse_std}
\begin{tabular}{l c c c c c c}
\toprule
  & \multicolumn{3}{c}{\textbf{CI}} & \multicolumn{3}{c}{\textbf{CD}} \\
\cmidrule(lr){2-4}\cmidrule(lr){5-7}
Dataset & TSM\textsubscript{CI} & DLin. & CycNet & TSM & SimTM & iTr. \\
\midrule
\multicolumn{7}{l}{\textit{Standard MTSF}} \\
ETTh1 & 0.001 & 0.012 & $<$0.001 & 0.009 & $<$0.001 & 0.003 \\
ETTh2 & 0.001 & 0.012 & 0.003 & 0.003 & 0.001 & 0.001 \\
ETTm1 & $<$0.001 & 0.006 & $<$0.001 & 0.002 & 0.003 & 0.004 \\
ETTm2 & 0.001 & 0.003 & 0.001 & 0.002 & 0.001 & 0.001 \\
Weather & 0.001 & 0.002 & 0.001 & 0.002 & 0.001 & 0.002 \\
Electricity & $<$0.001 & $<$0.001 & $<$0.001 & 0.001 & 0.009 & $<$0.001 \\
Traffic & $<$0.001 & $<$0.001 & $<$0.001 & 0.001 & 0.007 & 0.018 \\
ExchangeRate & $<$0.001 & 0.003 & 0.003 & 0.001 & 0.001 & 0.004 \\
ILI & 0.062 & 0.018 & 0.042 & 0.087 & 0.006 & 0.223 \\
SolarAL & $<$0.001 & $<$0.001 & 0.001 & 0.001 & 0.004 & 0.001 \\
\midrule
\multicolumn{7}{l}{\textit{Spatio-temporal (traffic)}} \\
METR-LA & 0.002 & 0.002 & 0.004 & 0.011 & 0.003 & 0.009 \\
PEMS-BAY & $<$0.001 & $<$0.001 & 0.005 & 0.002 & 0.001 & 0.003 \\
\midrule
\multicolumn{7}{l}{\textit{Coupled}} \\
AirQuality & 0.001 & $<$0.001 & $<$0.001 & 0.003 & 0.002 & 0.003 \\
AQShunyi & $<$0.001 & $<$0.001 & $<$0.001 & 0.005 & 0.004 & 0.001 \\
CzeLan & $<$0.001 & $<$0.001 & 0.001 & 0.001 & 0.003 & 0.002 \\
ZafNoo & 0.004 & 0.002 & 0.001 & 0.004 & 0.006 & 0.002 \\
Bavaria-Iller & $<$0.001 & 0.001 & $<$0.001 & 0.001 & $<$0.001 & $<$0.001 \\
Bavaria-Donau & $<$0.001 & 0.001 & 0.003 & 0.001 & 0.001 & 0.002 \\
WaterQuality-Darwin & 0.002 & 0.001 & 0.004 & 0.004 & 0.001 & 0.002 \\
Current-Velocity & $<$0.001 & $<$0.001 & 0.001 & 0.001 & $<$0.001 & 0.002 \\
Wearable & $<$0.001 & $<$0.001 & 0.001 & 0.001 & $<$0.001 & $<$0.001 \\
HouseholdPower & $<$0.001 & $<$0.001 & 0.001 & 0.001 & 0.001 & 0.002 \\
\bottomrule
\end{tabular}
\end{table*}

\begin{table*}[t]
\centering\small
\caption{Across-seed std of the horizon-averaged test MAE (seeds 2021/2025/2026), reported to gauge significance. Columns: TSM\textsubscript{CI}=TSMixer\textsubscript{CI}, DLin.=DLinear, CycNet=CycleNet (CI), and TSM=TSMixer, SimTM=SimpleTM, iTr.=iTransformer (CD).}
\label{tab:seed_mae_std}
\begin{tabular}{l c c c c c c}
\toprule
  & \multicolumn{3}{c}{\textbf{CI}} & \multicolumn{3}{c}{\textbf{CD}} \\
\cmidrule(lr){2-4}\cmidrule(lr){5-7}
Dataset & TSM\textsubscript{CI} & DLin. & CycNet & TSM & SimTM & iTr. \\
\midrule
\multicolumn{7}{l}{\textit{Standard MTSF}} \\
ETTh1 & 0.001 & 0.007 & $<$0.001 & 0.004 & $<$0.001 & 0.002 \\
ETTh2 & $<$0.001 & 0.006 & 0.002 & 0.001 & 0.001 & 0.001 \\
ETTm1 & $<$0.001 & 0.006 & $<$0.001 & 0.002 & 0.001 & 0.003 \\
ETTm2 & $<$0.001 & 0.003 & $<$0.001 & 0.001 & 0.001 & $<$0.001 \\
Weather & 0.001 & 0.002 & 0.001 & $<$0.001 & 0.001 & 0.002 \\
Electricity & $<$0.001 & $<$0.001 & $<$0.001 & 0.001 & 0.010 & $<$0.001 \\
Traffic & $<$0.001 & $<$0.001 & $<$0.001 & 0.002 & 0.006 & 0.013 \\
ExchangeRate & $<$0.001 & 0.003 & 0.002 & $<$0.001 & 0.001 & 0.002 \\
ILI & 0.021 & 0.004 & 0.019 & 0.016 & 0.006 & 0.059 \\
SolarAL & $<$0.001 & 0.001 & $<$0.001 & 0.001 & 0.005 & $<$0.001 \\
\midrule
\multicolumn{7}{l}{\textit{Spatio-temporal (traffic)}} \\
METR-LA & $<$0.001 & 0.001 & 0.004 & 0.002 & 0.002 & 0.004 \\
PEMS-BAY & $<$0.001 & $<$0.001 & $<$0.001 & 0.002 & 0.001 & 0.001 \\
\midrule
\multicolumn{7}{l}{\textit{Coupled}} \\
AirQuality & 0.001 & $<$0.001 & $<$0.001 & 0.001 & 0.001 & 0.002 \\
AQShunyi & $<$0.001 & $<$0.001 & $<$0.001 & 0.002 & 0.001 & 0.001 \\
CzeLan & $<$0.001 & 0.001 & 0.002 & 0.001 & 0.001 & 0.001 \\
ZafNoo & 0.002 & 0.001 & $<$0.001 & 0.002 & 0.003 & $<$0.001 \\
Bavaria-Iller & 0.001 & 0.002 & $<$0.001 & 0.001 & $<$0.001 & 0.001 \\
Bavaria-Donau & $<$0.001 & 0.002 & 0.003 & 0.001 & 0.001 & 0.002 \\
WaterQuality-Darwin & 0.002 & $<$0.001 & 0.001 & $<$0.001 & 0.003 & 0.001 \\
Current-Velocity & $<$0.001 & $<$0.001 & 0.001 & 0.001 & $<$0.001 & 0.002 \\
Wearable & $<$0.001 & $<$0.001 & $<$0.001 & $<$0.001 & $<$0.001 & $<$0.001 \\
HouseholdPower & $<$0.001 & $<$0.001 & $<$0.001 & 0.001 & $<$0.001 & 0.002 \\
\bottomrule
\end{tabular}
\end{table*}

\section{Per-horizon Results}
\label{sec:all-horizons-results}

In this section, we detail based on the horizons discussed in \ref{sec:datasets-details} the results for all models on all datasets. For better organization of results, we split them into three seperate tables, Table~\ref{tab:perh_standard}, Table~\ref{tab:perh_spatiotemporal}, and Table~\ref{tab:perh_proposed} showing the results for the standard, spatio temporal and the coupled datasets respectively.

\begin{table*}[p]
\centering
\caption{Per-horizon test MSE/MAE on the standard MTSF datasets (mean over seeds 2021/2025/2026). Horizon averages are in Tables~\ref{tab:seed_mse_mean}/\ref{tab:seed_mae_mean}. Best per row and metric \textbf{bold}, second \underline{underlined} (ties at three decimals all bold), and 1\textsuperscript{st} count = wins. Shading and abbreviations as in Table~\ref{tab:seed_mse_mean}.}
\label{tab:perh_standard}
\setlength{\tabcolsep}{2.2pt}
\resizebox{0.8\textwidth}{!}{%
\begin{tabular}{l c >{\columncolor{ciband}}c>{\columncolor{ciband}}c >{\columncolor{ciband}}c>{\columncolor{ciband}}c >{\columncolor{ciband}}c>{\columncolor{ciband}}c >{\columncolor{cdband}}c>{\columncolor{cdband}}c >{\columncolor{cdband}}c>{\columncolor{cdband}}c >{\columncolor{cdband}}c>{\columncolor{cdband}}c}
\toprule
  &   & \multicolumn{6}{c}{\cellcolor{ciband}\textbf{CI}} & \multicolumn{6}{c}{\cellcolor{cdband}\textbf{CD}} \\
\cmidrule(lr){3-8}\cmidrule(lr){9-14}
Dataset & $H$ & \multicolumn{2}{c}{TSM\textsubscript{CI}} & \multicolumn{2}{c}{DLin.} & \multicolumn{2}{c}{CycNet} & \multicolumn{2}{c}{TSM} & \multicolumn{2}{c}{SimTM} & \multicolumn{2}{c}{iTr.} \\
 & & MSE & MAE & MSE & MAE & MSE & MAE & MSE & MAE & MSE & MAE & MSE & MAE \\
\midrule
ETTh1 & 96 & 0.378 & \textbf{0.395} & \underline{0.374} & \underline{0.396} & \textbf{0.372} & \textbf{0.395} & 0.416 & 0.417 & 0.379 & 0.397 & 0.397 & 0.413 \\
 & 192 & \underline{0.419} & \underline{0.426} & 0.453 & 0.448 & \textbf{0.401} & \textbf{0.413} & 0.456 & 0.447 & 0.426 & 0.435 & 0.454 & 0.454 \\
 & 336 & 0.445 & \underline{0.443} & 0.469 & 0.457 & \textbf{0.431} & \textbf{0.430} & 0.460 & 0.455 & \underline{0.444} & 0.449 & 0.460 & 0.461 \\
 & 720 & 0.490 & 0.486 & 0.528 & 0.527 & \textbf{0.448} & \textbf{0.463} & 0.658 & 0.574 & \underline{0.457} & \underline{0.478} & 0.496 & 0.507 \\
\cmidrule(lr){1-14}
ETTh2 & 96 & 0.293 & \underline{0.343} & \underline{0.283} & 0.347 & \textbf{0.281} & \textbf{0.338} & 0.311 & 0.355 & 0.296 & 0.349 & 0.293 & 0.349 \\
 & 192 & \underline{0.376} & \underline{0.396} & 0.380 & 0.414 & \textbf{0.361} & \textbf{0.393} & 0.415 & 0.416 & 0.378 & \underline{0.396} & 0.390 & 0.399 \\
 & 336 & 0.439 & 0.441 & 0.421 & 0.446 & \textbf{0.403} & \textbf{0.426} & 0.484 & 0.463 & \underline{0.414} & \underline{0.431} & 0.420 & 0.433 \\
 & 720 & 0.453 & 0.454 & 0.680 & 0.588 & 0.436 & 0.449 & 0.436 & 0.455 & \underline{0.427} & \underline{0.447} & \textbf{0.426} & \textbf{0.445} \\
\cmidrule(lr){1-14}
ETTm1 & 96 & 0.289 & \underline{0.339} & \textbf{0.286} & \textbf{0.335} & 0.306 & 0.354 & \underline{0.288} & 0.342 & 0.289 & 0.343 & 0.304 & 0.353 \\
 & 192 & \underline{0.330} & \underline{0.366} & 0.335 & \textbf{0.364} & 0.337 & 0.375 & 0.331 & 0.368 & \textbf{0.329} & 0.368 & 0.355 & 0.389 \\
 & 336 & \textbf{0.359} & 0.389 & 0.369 & \textbf{0.384} & \underline{0.365} & \underline{0.386} & 0.401 & 0.408 & 0.369 & 0.391 & 0.374 & 0.396 \\
 & 720 & \underline{0.430} & \underline{0.427} & 0.447 & 0.443 & \textbf{0.409} & \textbf{0.410} & 0.456 & 0.448 & 0.442 & 0.432 & 0.436 & 0.436 \\
\cmidrule(lr){1-14}
ETTm2 & 96 & 0.170 & \underline{0.254} & \underline{0.163} & 0.257 & \textbf{0.160} & \textbf{0.250} & 0.171 & 0.256 & 0.170 & 0.256 & 0.180 & 0.269 \\
 & 192 & 0.236 & 0.301 & \underline{0.220} & 0.300 & \textbf{0.215} & \textbf{0.289} & 0.234 & 0.305 & 0.234 & \underline{0.299} & 0.244 & 0.312 \\
 & 336 & 0.291 & \underline{0.336} & \underline{0.283} & 0.347 & \textbf{0.269} & \textbf{0.328} & 0.289 & 0.340 & 0.284 & 0.338 & 0.288 & 0.341 \\
 & 720 & \underline{0.366} & \underline{0.390} & 0.385 & 0.415 & \textbf{0.353} & \textbf{0.383} & 0.390 & 0.400 & 0.371 & 0.391 & 0.380 & 0.398 \\
\cmidrule(lr){1-14}
Weather & 96 & 0.150 & 0.200 & \textbf{0.145} & 0.211 & \underline{0.148} & \underline{0.199} & 0.149 & 0.205 & 0.150 & \textbf{0.198} & 0.159 & 0.206 \\
 & 192 & 0.200 & 0.252 & \textbf{0.185} & \underline{0.248} & 0.206 & 0.257 & \underline{0.191} & \textbf{0.243} & 0.197 & \textbf{0.243} & 0.204 & 0.254 \\
 & 336 & 0.246 & \underline{0.286} & \textbf{0.237} & 0.294 & 0.261 & 0.298 & 0.251 & 0.289 & \underline{0.245} & \textbf{0.284} & 0.264 & 0.299 \\
 & 720 & 0.334 & 0.346 & \underline{0.322} & 0.365 & \textbf{0.313} & \textbf{0.333} & 0.348 & 0.350 & 0.347 & 0.356 & 0.323 & \underline{0.341} \\
\cmidrule(lr){1-14}
Electricity & 96 & \textbf{0.129} & \textbf{0.223} & 0.133 & 0.230 & \underline{0.131} & \underline{0.225} & 0.142 & 0.240 & \underline{0.131} & \underline{0.225} & 0.133 & 0.230 \\
 & 192 & \textbf{0.147} & \textbf{0.239} & \textbf{0.147} & 0.244 & \underline{0.148} & \underline{0.241} & 0.152 & 0.251 & 0.154 & 0.246 & 0.156 & 0.252 \\
 & 336 & \underline{0.162} & \textbf{0.256} & \underline{0.162} & \underline{0.261} & \textbf{0.159} & \textbf{0.256} & 0.169 & 0.273 & 0.172 & 0.269 & 0.169 & 0.265 \\
 & 720 & 0.195 & \underline{0.287} & 0.201 & 0.298 & 0.197 & \underline{0.287} & 0.202 & 0.298 & \textbf{0.189} & \textbf{0.285} & \underline{0.190} & \textbf{0.285} \\
\cmidrule(lr){1-14}
Traffic & 96 & \underline{0.359} & \textbf{0.251} & 0.385 & 0.269 & 0.381 & 0.266 & 0.380 & 0.267 & \textbf{0.355} & \underline{0.258} & 0.436 & 0.319 \\
 & 192 & \textbf{0.379} & \textbf{0.262} & 0.396 & 0.274 & 0.397 & \underline{0.269} & 0.426 & 0.319 & \underline{0.391} & 0.274 & 0.395 & 0.292 \\
 & 336 & \textbf{0.398} & \textbf{0.272} & 0.410 & 0.282 & \underline{0.405} & \underline{0.278} & 0.431 & 0.300 & 0.412 & 0.294 & 0.406 & 0.282 \\
 & 720 & \textbf{0.430} & \textbf{0.288} & 0.450 & 0.305 & 0.447 & 0.300 & 0.472 & 0.337 & 0.440 & 0.302 & \underline{0.433} & \underline{0.298} \\
\cmidrule(lr){1-14}
ExchangeRate & 30 & \textbf{0.028} & \textbf{0.113} & \textbf{0.028} & \textbf{0.113} & \underline{0.029} & \underline{0.114} & \underline{0.029} & 0.115 & \underline{0.029} & 0.115 & \underline{0.029} & \underline{0.114} \\
 & 90 & 0.077 & \underline{0.192} & \textbf{0.074} & \textbf{0.191} & 0.077 & 0.193 & 0.077 & 0.193 & \underline{0.076} & \underline{0.192} & 0.094 & 0.216 \\
 & 180 & \underline{0.160} & 0.284 & \textbf{0.144} & \textbf{0.282} & 0.171 & 0.292 & 0.162 & 0.285 & 0.162 & \underline{0.283} & 0.164 & 0.286 \\
 & 365 & \underline{0.344} & \underline{0.424} & \textbf{0.266} & \textbf{0.399} & 0.414 & 0.468 & 0.350 & 0.427 & 0.462 & 0.503 & 0.375 & 0.445 \\
\cmidrule(lr){1-14}
ILI & 24 & 1.857 & \underline{0.850} & 2.326 & 1.073 & 1.918 & 0.884 & 2.059 & 0.920 & \textbf{1.731} & \textbf{0.819} & \underline{1.839} & 0.879 \\
 & 36 & \textbf{1.756} & \textbf{0.878} & 2.294 & 1.082 & 1.824 & 0.905 & 1.997 & \underline{0.891} & \underline{1.813} & 0.898 & 2.091 & 0.919 \\
 & 48 & \underline{2.007} & \underline{0.979} & 2.545 & 1.131 & 2.113 & 1.006 & \textbf{1.920} & \textbf{0.915} & 2.156 & 0.989 & 3.216 & 1.281 \\
\cmidrule(lr){1-14}
SolarAL & 96 & \textbf{0.173} & \textbf{0.225} & 0.222 & 0.294 & 0.196 & 0.236 & 0.196 & 0.245 & 0.183 & \underline{0.232} & \underline{0.177} & 0.242 \\
 & 192 & 0.193 & \textbf{0.233} & 0.211 & 0.272 & 0.199 & \underline{0.253} & \textbf{0.187} & 0.258 & 0.197 & 0.259 & \underline{0.192} & 0.259 \\
 & 336 & \textbf{0.198} & \textbf{0.245} & 0.228 & 0.286 & 0.203 & \underline{0.257} & \underline{0.202} & 0.269 & 0.210 & 0.266 & 0.208 & 0.274 \\
 & 720 & \textbf{0.207} & \textbf{0.257} & 0.236 & 0.294 & 0.211 & \underline{0.265} & 0.216 & 0.273 & \underline{0.210} & 0.270 & 0.212 & 0.269 \\
\midrule
\multicolumn{2}{l}{1\textsuperscript{st} count} & 11 & 14 & 9 & 7 & 14 & 14 & 2 & 2 & 4 & 5 & 1 & 2 \\
\bottomrule
\end{tabular}}
\end{table*}

\begin{table*}[t]
\centering
\caption{Per-horizon test MSE/MAE on the spatio-temporal (traffic) datasets (mean over seeds 2021/2025/2026). Horizon averages are in Tables~\ref{tab:seed_mse_mean}/\ref{tab:seed_mae_mean}. Best per row and metric \textbf{bold}, second \underline{underlined} (ties at three decimals all bold), and 1\textsuperscript{st} count = wins. Shading and abbreviations as in Table~\ref{tab:seed_mse_mean}.}
\label{tab:perh_spatiotemporal}
\setlength{\tabcolsep}{2.2pt}
\resizebox{0.8\textwidth}{!}{%
\begin{tabular}{l c >{\columncolor{ciband}}c>{\columncolor{ciband}}c >{\columncolor{ciband}}c>{\columncolor{ciband}}c >{\columncolor{ciband}}c>{\columncolor{ciband}}c >{\columncolor{cdband}}c>{\columncolor{cdband}}c >{\columncolor{cdband}}c>{\columncolor{cdband}}c >{\columncolor{cdband}}c>{\columncolor{cdband}}c}
\toprule
  &   & \multicolumn{6}{c}{\cellcolor{ciband}\textbf{CI}} & \multicolumn{6}{c}{\cellcolor{cdband}\textbf{CD}} \\
\cmidrule(lr){3-8}\cmidrule(lr){9-14}
Dataset & $H$ & \multicolumn{2}{c}{TSM\textsubscript{CI}} & \multicolumn{2}{c}{DLin.} & \multicolumn{2}{c}{CycNet} & \multicolumn{2}{c}{TSM} & \multicolumn{2}{c}{SimTM} & \multicolumn{2}{c}{iTr.} \\
 & & MSE & MAE & MSE & MAE & MSE & MAE & MSE & MAE & MSE & MAE & MSE & MAE \\
\midrule
METR-LA & 12 & 0.431 & 0.337 & 0.430 & 0.345 & 0.433 & 0.344 & \textbf{0.422} & \textbf{0.321} & \underline{0.424} & \underline{0.322} & 0.453 & 0.355 \\
 & 48 & 0.817 & 0.522 & 0.838 & 0.569 & \textbf{0.806} & \underline{0.521} & \underline{0.810} & 0.543 & 0.892 & \textbf{0.506} & 0.881 & 0.546 \\
 & 144 & 1.162 & 0.659 & \textbf{1.106} & 0.699 & \underline{1.156} & \textbf{0.645} & 1.170 & \underline{0.656} & 1.178 & 0.662 & 1.346 & 0.695 \\
 & 288 & \underline{1.255} & 0.701 & \textbf{1.217} & 0.736 & 1.281 & \underline{0.693} & 1.319 & \textbf{0.678} & 1.310 & 0.694 & 1.400 & 0.725 \\
\cmidrule(lr){1-14}
PEMS-BAY & 12 & 0.294 & 0.260 & 0.317 & 0.267 & 0.273 & 0.245 & \textbf{0.247} & \underline{0.225} & 0.264 & 0.233 & \underline{0.252} & \textbf{0.224} \\
 & 48 & 0.465 & 0.324 & 0.799 & 0.497 & 0.508 & 0.343 & 0.483 & 0.326 & \underline{0.441} & \underline{0.310} & \textbf{0.435} & \textbf{0.293} \\
 & 144 & 0.600 & 0.368 & 0.656 & 0.414 & 0.641 & 0.394 & 0.547 & 0.353 & \textbf{0.468} & \textbf{0.318} & \underline{0.534} & \underline{0.335} \\
 & 288 & 0.646 & 0.401 & 0.678 & 0.425 & 0.689 & 0.427 & 0.601 & 0.387 & \textbf{0.520} & \textbf{0.329} & \underline{0.564} & \underline{0.349} \\
\midrule
\multicolumn{2}{l}{1\textsuperscript{st} count} & 0 & 0 & 2 & 0 & 1 & 1 & 2 & 2 & 2 & 3 & 1 & 2 \\
\bottomrule
\end{tabular}}
\end{table*}

\begin{table*}[p]
\centering
\caption{Per-horizon test MSE/MAE on the coupled datasets (mean over seeds 2021/2025/2026). Horizon averages are in Tables~\ref{tab:seed_mse_mean}/\ref{tab:seed_mae_mean}. Best per row and metric \textbf{bold}, second \underline{underlined} (ties at three decimals all bold), and 1\textsuperscript{st} count = wins. Shading and abbreviations as in Table~\ref{tab:seed_mse_mean}.}
\label{tab:perh_proposed}
\setlength{\tabcolsep}{2.2pt}
\resizebox{0.8\textwidth}{!}{%
\begin{tabular}{l c >{\columncolor{ciband}}c>{\columncolor{ciband}}c >{\columncolor{ciband}}c>{\columncolor{ciband}}c >{\columncolor{ciband}}c>{\columncolor{ciband}}c >{\columncolor{cdband}}c>{\columncolor{cdband}}c >{\columncolor{cdband}}c>{\columncolor{cdband}}c >{\columncolor{cdband}}c>{\columncolor{cdband}}c}
\toprule
  &   & \multicolumn{6}{c}{\cellcolor{ciband}\textbf{CI}} & \multicolumn{6}{c}{\cellcolor{cdband}\textbf{CD}} \\
\cmidrule(lr){3-8}\cmidrule(lr){9-14}
Dataset & $H$ & \multicolumn{2}{c}{TSM\textsubscript{CI}} & \multicolumn{2}{c}{DLin.} & \multicolumn{2}{c}{CycNet} & \multicolumn{2}{c}{TSM} & \multicolumn{2}{c}{SimTM} & \multicolumn{2}{c}{iTr.} \\
 & & MSE & MAE & MSE & MAE & MSE & MAE & MSE & MAE & MSE & MAE & MSE & MAE \\
\midrule
AirQuality & 96 & 0.693 & 0.476 & \underline{0.678} & 0.474 & 0.683 & \textbf{0.467} & \textbf{0.674} & \underline{0.469} & 0.687 & 0.473 & 0.691 & 0.474 \\
 & 192 & 0.737 & 0.494 & \textbf{0.720} & 0.496 & \underline{0.726} & \textbf{0.487} & 0.769 & 0.501 & 0.730 & \underline{0.490} & 0.727 & \underline{0.490} \\
 & 336 & 0.756 & 0.506 & \underline{0.748} & 0.515 & 0.750 & \textbf{0.499} & \textbf{0.740} & 0.502 & 0.752 & 0.502 & 0.751 & \underline{0.501} \\
 & 720 & 0.815 & 0.530 & \underline{0.805} & 0.532 & 0.817 & 0.527 & \textbf{0.773} & \textbf{0.504} & 0.839 & \underline{0.521} & 0.822 & 0.532 \\
\cmidrule(lr){1-14}
AQShunyi & 96 & 0.639 & 0.484 & \textbf{0.621} & \underline{0.480} & \underline{0.628} & \textbf{0.476} & 0.652 & 0.491 & 0.630 & \textbf{0.476} & 0.678 & 0.491 \\
 & 192 & 0.676 & 0.502 & \underline{0.659} & 0.502 & 0.665 & \textbf{0.494} & 0.667 & \underline{0.499} & \textbf{0.652} & \textbf{0.494} & 0.670 & 0.500 \\
 & 336 & 0.687 & 0.509 & \underline{0.681} & 0.516 & 0.694 & 0.516 & \underline{0.681} & \underline{0.507} & \textbf{0.670} & \textbf{0.501} & 0.699 & 0.514 \\
 & 720 & 0.733 & 0.533 & 0.725 & 0.535 & 0.743 & 0.533 & \underline{0.707} & \underline{0.520} & \textbf{0.705} & \textbf{0.516} & 0.742 & 0.534 \\
\cmidrule(lr){1-14}
CzeLan & 96 & 0.172 & \textbf{0.219} & \underline{0.167} & 0.227 & 0.178 & 0.236 & \textbf{0.162} & \underline{0.221} & 0.169 & 0.228 & 0.171 & 0.235 \\
 & 192 & 0.200 & \textbf{0.246} & \underline{0.195} & 0.253 & 0.202 & \textbf{0.246} & \textbf{0.193} & \underline{0.247} & 0.204 & 0.252 & 0.234 & 0.289 \\
 & 336 & \textbf{0.222} & \textbf{0.265} & \underline{0.225} & 0.284 & 0.231 & \underline{0.276} & 0.228 & 0.281 & 0.236 & 0.284 & 0.242 & 0.284 \\
 & 720 & 0.253 & 0.296 & 0.286 & 0.343 & 0.255 & 0.299 & \underline{0.239} & \underline{0.287} & \textbf{0.234} & \textbf{0.284} & 0.254 & 0.308 \\
\cmidrule(lr){1-14}
ZafNoo & 96 & 0.441 & \underline{0.405} & \textbf{0.419} & \textbf{0.393} & 0.447 & 0.408 & \underline{0.431} & \underline{0.405} & 0.445 & 0.410 & 0.457 & 0.419 \\
 & 192 & 0.506 & 0.453 & \textbf{0.477} & \textbf{0.433} & 0.511 & \underline{0.446} & \underline{0.496} & 0.452 & 0.514 & 0.463 & 0.507 & 0.449 \\
 & 336 & 0.544 & 0.479 & \textbf{0.508} & \textbf{0.447} & 0.560 & 0.476 & \underline{0.538} & 0.477 & 0.553 & 0.494 & 0.552 & \underline{0.474} \\
 & 720 & 0.603 & 0.513 & \textbf{0.541} & \textbf{0.485} & 0.594 & 0.512 & \underline{0.589} & \underline{0.503} & 0.604 & 0.512 & 0.641 & 0.545 \\
\cmidrule(lr){1-14}
Bavaria-Iller & 24 & 0.040 & 0.079 & 0.041 & 0.082 & 0.047 & 0.083 & \textbf{0.035} & \underline{0.077} & \underline{0.036} & \textbf{0.072} & 0.043 & 0.080 \\
 & 96 & 0.172 & 0.179 & 0.171 & 0.201 & 0.174 & 0.182 & \textbf{0.154} & \textbf{0.171} & \underline{0.156} & \underline{0.174} & 0.172 & 0.179 \\
 & 336 & 0.344 & 0.304 & \textbf{0.315} & 0.339 & 0.336 & \textbf{0.300} & 0.348 & 0.304 & \underline{0.318} & \underline{0.301} & 0.342 & 0.302 \\
\cmidrule(lr){1-14}
Bavaria-Donau & 24 & 0.012 & 0.060 & 0.011 & 0.058 & 0.011 & 0.059 & \textbf{0.009} & \textbf{0.051} & \underline{0.010} & \underline{0.055} & 0.011 & 0.057 \\
 & 96 & 0.058 & 0.142 & 0.066 & 0.155 & 0.073 & 0.166 & \textbf{0.050} & \textbf{0.130} & \underline{0.054} & \underline{0.131} & 0.064 & 0.153 \\
 & 336 & 0.221 & 0.306 & 0.210 & 0.291 & 0.210 & 0.295 & \underline{0.200} & \underline{0.284} & \textbf{0.194} & \textbf{0.281} & 0.224 & 0.305 \\
\cmidrule(lr){1-14}
WaterQuality-Darwin & 24 & \underline{0.222} & \underline{0.169} & 0.230 & 0.171 & 0.229 & 0.170 & \textbf{0.218} & \textbf{0.165} & 0.230 & 0.171 & 0.229 & 0.173 \\
 & 96 & \underline{0.317} & 0.207 & 0.332 & 0.209 & \textbf{0.314} & 0.209 & \underline{0.317} & \underline{0.205} & 0.321 & \textbf{0.204} & 0.332 & 0.208 \\
\cmidrule(lr){1-14}
Current-Velocity & 24 & 0.230 & \underline{0.243} & 0.237 & 0.248 & 0.237 & 0.248 & \textbf{0.223} & \textbf{0.236} & \underline{0.224} & \textbf{0.236} & 0.234 & 0.248 \\
 & 96 & 0.295 & 0.295 & 0.291 & 0.293 & 0.302 & 0.299 & \underline{0.287} & \underline{0.289} & \textbf{0.285} & \textbf{0.288} & 0.296 & 0.295 \\
\cmidrule(lr){1-14}
Wearable & 12 & 0.647 & 0.576 & \textbf{0.638} & \underline{0.571} & 0.734 & 0.576 & \underline{0.639} & \underline{0.571} & 0.723 & \textbf{0.570} & 0.727 & 0.574 \\
 & 48 & 0.673 & \underline{0.591} & \textbf{0.669} & \underline{0.591} & 0.778 & 0.599 & \underline{0.672} & 0.592 & 0.758 & \textbf{0.590} & 0.761 & 0.593 \\
 & 144 & 0.678 & 0.596 & \underline{0.676} & \underline{0.595} & 0.802 & 0.605 & \textbf{0.671} & \textbf{0.593} & 0.778 & 0.598 & 0.779 & 0.599 \\
 & 288 & \underline{0.685} & \textbf{0.604} & 0.687 & \underline{0.605} & 0.809 & 0.615 & \textbf{0.683} & \underline{0.605} & 0.754 & 0.607 & 0.754 & 0.608 \\
\cmidrule(lr){1-14}
HouseholdPower & 96 & 0.629 & 0.493 & 0.623 & 0.501 & 0.623 & 0.493 & 0.625 & 0.494 & \underline{0.622} & \underline{0.490} & \textbf{0.621} & \textbf{0.488} \\
 & 192 & 0.643 & 0.505 & \underline{0.637} & 0.510 & \textbf{0.635} & 0.502 & 0.644 & 0.508 & 0.642 & \textbf{0.498} & 0.641 & \underline{0.501} \\
 & 336 & 0.663 & 0.515 & \textbf{0.650} & 0.518 & 0.652 & 0.509 & 0.653 & 0.512 & 0.654 & \textbf{0.507} & \underline{0.651} & \underline{0.508} \\
 & 672 & 0.674 & 0.523 & \textbf{0.666} & 0.528 & 0.674 & \textbf{0.516} & 0.671 & 0.520 & 0.672 & 0.521 & \underline{0.669} & \underline{0.518} \\
\midrule
\multicolumn{2}{l}{1\textsuperscript{st} count} & 1 & 4 & 11 & 4 & 2 & 8 & 13 & 7 & 6 & 14 & 1 & 1 \\
\bottomrule
\end{tabular}}
\end{table*}

\section{Synthetic Data Generation Description}
\label{sec:synthetic-data-generation-details}
Each synthetic dataset has four channels $c_0,\dots,c_3$ observed over $T=20{,}000$ hourly timesteps.
Channels $c_1,c_2,c_3$ are independent sources. The target $c_0$ receives the planted coupling.
All six datasets share the same four channel bases, drawn from the same random streams, and
differ \emph{only} in the edges planted into $c_0$. Generation proceeds in the five steps below.
Throughout, $z[\cdot]$ denotes standardization over the full sample,
\begin{equation}
  z[v](t) = \frac{v(t)-\bar v}{\hat\sigma_v},\qquad
  \bar v = \tfrac1T\textstyle\sum_{t} v(t),\quad
  \hat\sigma_v^2 = \tfrac1T\textstyle\sum_{t}\bigl(v(t)-\bar v\bigr)^2 ,
  \label{eq:synth_z}
\end{equation}
so every weight below is a variance share.

\paragraph{Step 1: noise.}
Each channel $c_i$ has its own pseudo-random generator, seeded from the dataset seed ($21$) and
the channel name, so a channel's draws do not depend on the order in which channels are
generated. From it we draw Gaussian innovations and one seasonal phase:
\begin{equation}
  \varepsilon_i(t)\overset{\text{i.i.d.}}{\sim}\mathcal N(0,1),\quad t=-B,\dots,T-1,\ B=200;
  \qquad
  \phi_i\sim\mathcal U[0,2\pi).
  \label{eq:synth_noise}
\end{equation}
Innovations of different channels are independent, so the bases built from them are mutually
independent processes.

\paragraph{Step 2: stochastic AR($p$) core.}
The core of $c_i$ is an autoregression of order $p_i$, started at zero and run from $t=-B$, with
the first $B$ samples discarded as burn-in:
\begin{equation}
  \eta_i(t) = \sum_{k=1}^{p_i} a_{i,k}\,\eta_i(t-k) + \varepsilon_i(t),
  \qquad
  a_{i,k} = \rho_i\,\frac{e^{-(k-1)/\tau_i}}{\sum_{j=1}^{p_i}e^{-(j-1)/\tau_i}},
  \quad \tau_i=\tfrac{p_i}{2}.
  \label{eq:synth_ar}
\end{equation}
The coefficients decay exponentially and sum to the persistence $\rho_i<1$. Since
$a_{i,k}>0$, for $|x|\le1$ we have $\bigl|\sum_k a_{i,k}x^k\bigr|\le\rho_i<1$, so the
characteristic polynomial $1-\sum_k a_{i,k}x^k$ has no root in the closed unit disc and every
core is stationary. The orders and persistences are
$(p_i,\rho_i) = (4,0.7),\,(5,0.8),\,(3,0.6),\,(4,0.7)$ for $i=0,\dots,3$. For example,
$(a_{0,1},\dots,a_{0,4}) = (0.319,\,0.193,\,0.117,\,0.071)$.

\paragraph{Step 3: seasonal and trend components.}
Each channel carries one random-phase sinusoid and, optionally, a linear trend:
\begin{equation}
  s_i(t) = \sin\!\Bigl(\frac{2\pi t}{P_i}+\phi_i\Bigr),
  \qquad
  g_i(t) = \beta_i\,t .
  \label{eq:synth_season}
\end{equation}
The periods are distinct across channels, $(P_0,\dots,P_3)=(24,168,8,12)$, so no two channels share
a frequency. A shared frequency would create an artificial periodic cross-correlation between
channels that are not coupled. For the same reason, the trend is disabled in the suite
($\beta_i=0$). Two independent trending series are correlated in levels (spurious regression)
and are non-stationary, which would contaminate the ground truth.

\paragraph{Step 4: channel base.}
The components are combined \emph{additively} at an explicit variance split
$\pi_i=(\pi_{i,\eta},\pi_{i,s},\pi_{i,g})$, with $\sum_m\pi_{i,m}=1$:
\begin{equation}
  u_i = z\Bigl[\sqrt{\pi_{i,\eta}}\;z[\eta_i] + \sqrt{\pi_{i,s}}\;z[s_i] + \sqrt{\pi_{i,g}}\;z[g_i]\Bigr].
  \label{eq:synth_base}
\end{equation}
The seasonal and trend terms are added to the output of the AR recursion, not fed through
it, so the AR polynomial does not filter the season. Because the components are standardized
before weighting, $\pi_{i,m}$ is (up to finite-sample correlation) the fraction of the base's
variance explained by component $m$, whatever the AR persistence or the sinusoid amplitude. We use
$\pi_{i,g}=0$ and
$(\pi_{i,\eta},\pi_{i,s}) = (0.75,0.25),\,(0.80,0.20),\,(0.85,0.15),\,(0.80,0.20)$ for
$i=0,\dots,3$. With $z[s_i]=\sqrt2\,s_i$, the target's base is thus
\begin{equation}
  u_0(t) \approx 0.866\;z[\eta_0](t) + 0.707\,\sin\!\Bigl(\frac{2\pi t}{24}+\phi_0\Bigr).
  \label{eq:synth_u0}
\end{equation}
The sources have no incoming edges and are final at this point: $c_i=z[u_i]$ for $i=1,2,3$.

\paragraph{Step 5: planting the coupling.}
The target mixes its own base with the standardized contributions of its incoming edges
$e\in\mathcal E_0$:
\begin{equation}
  c_0 = z\Bigl[\sqrt{\omega}\;u_0 + \sum_{e\in\mathcal E_0}\sqrt{w_e}\;z\bigl[f_e(c_{\mathrm{src}_e})\bigr]\Bigr],
  \qquad \omega+\sum_{e}w_e=1 .
  \label{eq:synth_mix}
\end{equation}
$\omega$ is the own share and $w_e$ the strength of edge $e$. $u_0$ is independent of every source,
and the edges of a dataset use independent sources, so the terms are uncorrelated and each explains
its nominal share of $\operatorname{Var}(c_0)$. The realized shares match to within $0.01$.
Standardizing each contribution before weighting ties the coupling strength to $w_e$ alone,
independent of the functional form of $f_e$. Each $f_e$ reads its source through the causal lag
operator $(\mathrm L^{\ell}x)(t)=x(t-\ell)$ for $t\ge\ell$, and $0$ otherwise, and is one of
\begin{equation}
  f(x)(t) \in \Bigl\{\, x(t),\;\; x(t-\ell),\;\; x(t-\ell)^2,\;\; (c_a c_b)(t-\ell) \,\Bigr\},
  \label{eq:synth_f}
\end{equation}
i.e.\ contemporaneous linear, lagged linear, lagged non-linear, and multiplicative (joint)
coupling. The multiplicative coupling is a single edge with two sources $c_a,c_b$. With $\omega=0.6$
and $w_e=0.4$ for every single-edge dataset, the six targets are given below, as in Figure~\ref{fig:synth_graphs}. We omit the outer $z[\cdot]$ of Eq.~\eqref{eq:synth_mix}, which is close to the identity here, since the terms are uncorrelated with unit variance and their shares sum to one:
\begin{align}
  \ctype{independent}:\quad & c_0(t) = u_0(t), \nonumber\\
  \ctype{contemp-linear}:\quad & c_0(t) = \sqrt{0.6}\,u_0(t) + \sqrt{0.4}\;z[c_1(t)], \nonumber\\
  \ctype{lagged-linear}:\quad & c_0(t) = \sqrt{0.6}\,u_0(t) + \sqrt{0.4}\;z[c_1(t-24)], \nonumber\\
  \ctype{lagged-non-linear}:\quad & c_0(t) = \sqrt{0.6}\,u_0(t) + \sqrt{0.4}\;z[c_1(t-12)^2], \nonumber\\
  \ctype{multiplicative-joint}:\quad & c_0(t) = \sqrt{0.6}\,u_0(t) + \sqrt{0.4}\;z[(c_1c_2)(t-6)], \nonumber\\
  \ctype{additive-mixed}:\quad & c_0(t) = \sqrt{0.3}\,u_0(t) + \sqrt{0.4}\;z[c_1(t-24)] + \sqrt{0.3}\;z[c_2(t-8)^2]. \nonumber
\end{align}
For the linear contributions, the inner $z[\cdot]$ is the identity, because the sources are already standardized. In each
dataset, the sources without an edge into $c_0$ are independent controls. The four channels
$c_0,\dots,c_3$ are written as one CSV with an hourly timestamp and split chronologically
70/10/20 into train/validation/test, as for the single-series real-world datasets (Appendix~\ref{sec:datasets-details}).

\end{document}